\documentclass[11pt]{article}

\PassOptionsToPackage{dvipsnames,table}{xcolor}
\usepackage{acl}

\usepackage{times}
\usepackage{latexsym}

\usepackage[T1]{fontenc}
\usepackage[utf8]{inputenc}
\usepackage{microtype}
\usepackage{inconsolata}

\usepackage{graphicx}
\usepackage{multirow}

\usepackage{xspace}
\usepackage{amsmath}        
\usepackage{amsfonts}        
\usepackage{algorithm}
\usepackage{algpseudocode}
\usepackage{booktabs}       
\usepackage{enumitem}
\usepackage{placeins}       
\usepackage{cleveref}       
\newcommand{\err}[1]{{\scriptstyle \pm #1}}

\def\Eqref#1{Equation~\ref{#1}}

\title{Coresets Before Score Sets:\\ Evaluation-Unsupervised Prompt Subset Selection for LLM Benchmarks}

\author{
  \textbf{Jihan Yao\textsuperscript{1}\thanks{Equal contribution, random order.}},
  \textbf{Gantavya Bhatt\textsuperscript{1}\footnotemark[1]},
  \textbf{Arnav Das\textsuperscript{1}\footnotemark[1]},
  \textbf{Peter Jin\textsuperscript{6}},
  \textbf{Ke Bao\textsuperscript{5}},
  \textbf{Qiaolin Yu\textsuperscript{5}},
\\
  \textbf{Khushi Bhardwaj\textsuperscript{6}},
  \textbf{Chang Su\textsuperscript{3}},
  \textbf{Jialei Wang\textsuperscript{6}},
  \textbf{Yikai Zhu\textsuperscript{5}},
  \textbf{Sugam Devare\textsuperscript{6}},
\\
  \textbf{Damon Mosk-Aoyama\textsuperscript{6}},
  \textbf{Zhen Dong\textsuperscript{6}},
  \textbf{Venkat Krishna Srinivasan\textsuperscript{6}},
  \textbf{Yineng Zhang\textsuperscript{4}},
\\
  \textbf{Oleksii Kuchaiev\textsuperscript{6}},
  \textbf{Jiantao Jiao\textsuperscript{2,6}},
  \textbf{Banghua Zhu\textsuperscript{1,5}},
  \textbf{Jeff Bilmes\textsuperscript{1}}
\\
\\
  \textsuperscript{1}University of Washington, Seattle \quad
  \textsuperscript{2}University of California, Berkeley \quad
\\
  \textsuperscript{3}Oracle \quad
  \textsuperscript{4}Together AI \quad
  \textsuperscript{5}LMSYS \quad
  \textsuperscript{6}NVIDIA
\\
  \small{
    \textbf{Correspondence:}
    \texttt{jihany2@cs.washington.edu},
    \texttt{gbhatt2@uw.edu},
    \texttt{arnavmd2@uw.edu}
  }
}

\begin{document}
\maketitle

\begin{abstract}

We study LLM benchmark coreset selection: selecting a small subset of prompts
over multiple benchmarks whose induced model scores and rankings approximate
those obtained from the full benchmark suite. In evaluation-unsupervised
benchmark coreset selection (our approach), the selection algorithm uses no
model evaluation outcomes, and operates on a fine granularity by producing
subsets of prompts over multiple benchmarks rather than producing a
sub-collection of entire benchmarks. We use submodular subset selection, and we
develop and evaluate many different submodular functions for this purpose,
including determinantal point process (DPP) based approaches, submodular mutual
information functions, and facility location-based functions. On a new
large-scale suite of 35 heterogeneous benchmarks spanning five different
capability categories, 18 frontier LLMs, and over 61K prompts, we find that the
facility location (FL) function operating exclusively on inexpensive semantic prompt
embeddings preserves LLM scores better than twelve separate score-based and
diversity-based baselines, across a range of coreset budgets. Moreover, we show
our proposed objective is not limited to the evaluation-unsupervised regime: in
the setting where only a handful of whole benchmarks must be selected and a
large amount of model scores are available, the same objective matches or
outperforms state-of-the-art baselines on the MMLU and MTEB leaderboards, while
being substantially cheaper to compute. Together, our results suggest that
submodularity, in general, is a strong and reliable tool for benchmark
compression.

\end{abstract}

\section{Introduction}

LLM evaluation is a major but necessary cost in large language model (LLM)
development.  As models are deployed across increasingly specialized settings,
evaluation suites have expanded from a small number of general-purpose
benchmarks to large collections spanning knowledge, mathematics, coding,
instruction following, and agentic behavior~\citep{phan2025humanity,
wang2024mmlu, merrill2026terminal, pyatkin2026generalizing, yao2024tau}.
Typically, LLM evaluation is repeatedly performed throughout the full LLM
development cycle as new models are regularly created. However, repeatedly
running evaluations on the full suite of benchmarks is computationally expensive
and costly. Recently, a large body of work has shown that many benchmarks
contain substantial redundancy~\citep{datologyai2026datbench,
polo2024tinybenchmarks}, meaning that benchmark evaluation is naturally
inefficient, since different parts of different benchmarks might redundantly
evaluate the same model ability (also see Figure~\ref{fig:bench_similarity_matrix}). This has motivated a growing focus on
\emph{benchmark coresets}\footnote{We prefer the term ``benchmark coresets''
or benchmark ``subset selection'' to avoid overloading the term ``compression,''
which can also imply bit-level encoding, entropy, rate–distortion,
gzip/JPEG-style coding, and data distillation, as discussed in
~\cite{bilmes2022submodularity}.}: i.e., selecting a much smaller set of
examples or benchmarks (or portions thereof) that preserves the evaluation
results of a full evaluation suite~\citep{vivek2024anchor, kipnis2025metabench}.

In this paper we study LLM benchmark coreset selection: selecting a small subset
of benchmark prompts whose induced model scores and rankings approximate those
obtained from the complete benchmark suite. We distinguish between two variants
of this problem. 

In \textbf{evaluation-supervised benchmark coreset selection}\footnote{We
consider this a ``supervised'' rather than an ``unsupervised'' subset selection
procedure, consistent with previous terminology~\cite{wei2014unsupervised} in
data subset selection.}, the coreset algorithm may use model-by-prompt
evaluation outcomes, such as previous scores, correctness indicators, responses,
or item-response statistics, as supervision information available for subset
selection. This first case is more informationally rich than the second (below), but has two key
limitations. First, it is required that many models have already been evaluated
on the full set of benchmarks, and that their responses can be used to guide the
benchmark coreset algorithm~\citep{polo2024tinybenchmarks, vivek2024anchor,
kipnis2025metabench, smola2026submodular}. This assumption is reasonable for
established benchmarks such as MMLU~\cite{wang2024mmlu}. For a newly released
benchmark, however, such responses must be collected from scratch for every
model considered. As the ecosystem of benchmarks grows rapidly (as it should), benchmark
compression strategies that rely heavily on these responses can become
prohibitively expensive. The second limitation of the supervised approach is the
experimental scope: prior methods are typically evaluated on old benchmarks with
open leaderboards or large historical model responses, such as Open LLM
Leaderboard and MMLU~\citep{myrzakhan2024open, hendrycks2020measuring}. However,
recent work has shown that these benchmarks suffer from serious data
contamination and benchmark-specific overfitting
risks~\citep{deng2024investigating, xu2024benchmark}. As a result, conclusions
drawn from these benchmarks may carry biases from data contamination, and not
serve as true guidance for new benchmark development. This narrow experimental
scope also limits how fidelity is measured. Since prior studies are often
conducted on single or only a few benchmarks, they mainly evaluate whether
coresets preserves a single benchmark score or an overall average score. Such
aggregated fidelity can miss capability-wise failures: errors from different
capabilities may cancel each other out, making a compressed subset appear
faithful even when it distorts performance on specific capabilities. Therefore,
extensive experiments on recent, heterogeneous benchmark suite spanning multiple
capabilities are necessary.

In \textbf{evaluation-unsupervised benchmark coreset selection}, by contrast,
(the approach we introduce in this paper), the selection algorithm uses no
previous model evaluation outcomes and uses information only from the prompts
themselves. Indeed, we hypothesize that a useful benchmark coreset method should
operate in an \emph{evaluation-unsupervised} regime: selecting a subset from a
target benchmark should only utilize dataset-level observable information, such
as text, metadata, and optional capability labels, but with no, or at most very
limited, access to model responses or evaluation scores. 

Another distinct aspect of our approach is granularity: we produce coresets as
fine-grained subsets of prompts rather than the finding subsets of a set of benchmarks.
This allows us to extract non-redundant fractional portions of multiple
benchmarks to produce a highly tuned benchmark coreset that well represents all
of the original benchmarks better than the more coarse-grained approach of
selecting a subset of entire benchmarks.

Overall, in this paper, we study methods for the production of
\emph{evaluation-unsupervised benchmark coresets} at the fine granularity of
prompts. We evaluate a number of different strategies for this, many under the
submodular~\cite{bilmes2022submodularity} umbrella, a class of techniques that
has been successfully applied to many data subset selection problems in the
past, including document summarization, sensor placement, and active learning.
In particular, we evaluate many different submodular functions for this purpose,
including determinantal point process (DPP) based
approaches~\cite{kulesza2012determinantal}, submodular mutual information
functions~\cite{bilmes2022submodularity,iyer2021generalized,iyer2021submodular},
and facility location-based functions~\cite{bilmes2022submodularity}. Our
methods first construct a model-independent semantic representation for each
prompt in each benchmark by concatenating the prompt and ground-truth fields,
such as answers, constraints, and patches, and then embedding the concatenated
text. In the DPP case, the method uses these features directly, while in the
facility location case, it builds a sparse similarity matrix over items. In both
cases, it maximizes a submodular objective so that the selected subset covers
the full evaluation suite without requiring historical scores or responses.

As part of this work, and to help with evaluations, we introduce a new
large-scale score-evaluation suite of 35 heterogeneous benchmarks spanning five
capability categories, 18 frontier LLMs,  (including Claude--4.5 Sonnet, GPT--5, Gemini--2.5 Pro), and over 61K newly collected prompts. Unlike previous work, which mostly experiments on legacy benchmarks, our study examines coreset selection algorithms on a more generalizable and up-to-date scenario.

Across budgets $k \in [70,200]$,
facility location (FL) achieves the best overall and capability-wise fidelity,
with the lowest normalized overall mean relative error (0.008) and near-zero
normalized capability mean relative error, outperforming twelve score-based and
diversity-based baselines. Our analysis shows that this advantage comes from
coverage: score-based methods tend to select samples that are concentrated
within a single benchmark, while IRT-based embeddings learned from only 18
models are unstable. In contrast, semantic embeddings do not require model
responses or evaluation scores and provide a more reliable basis for selection.

We further find that the same FL objective remains effective beyond the
evaluation-unsupervised regime. We evaluate how well FL can
select a benchmark-level coreset that recovers held-out models' full scores on
the whole leaderboard when rich model evaluation scores are available. The
results show that FL is the strongest method across most budgets. In this
setting, where only a handful of whole benchmarks must be selected and many
model scores are available, the same objective matches or outperforms
state-of-the-art baselines on the MMLU and MTEB leaderboards, while being
substantially cheaper to compute. Together, our results suggest that
submodularity is a strong and reliable tool for benchmark coresets.


\section{Related Work}
\paragraph{Benchmark Coresets.}

Given the rapid growth in LLM benchmark datasets, a growing body of work has proposed various methods for reducing benchmarks to accelerate evaluation times. These methods generally select a subset of samples or subsets of benchmarks. At the sample level, existing methods mainly rely on observed model behavior: item-response theory~\citep{polo2024tinybenchmarks}, confidence-based anchor examples~\citep{vivek2024anchor}, model response matrices~\citep{kipnis2025metabench, huang2025minilongbench}, hidden-state representations~\citep{zhang2026learning}, or a small set of anchor LLMs combined with intrinsic sample features~\citep{saranathan2025sublime}. At the dataset level, existing methods select benchmark subsets using raw evaluation scores~\citep{subramani2025simbasimplifyingbenchmarkanalysis}, task-similarity measures from model response patterns~\citep{surkov2024vygotsky}, or covariance, imputation, and submodular objectives over benchmark scores~\citep{smola2026submodular}. Despite these differences, the majority of existing methods depend on model response or scores from aggregated public leaderboards, which are expensive to obtain and often sparse since not all leaderboards evaluate the same sets of models, limiting their applicability in new or private evaluation settings. In contrast, our method operates at the sample level in an unsupervised setting, selecting samples using only semantic embeddings and requiring very limited model evaluations, using them solely for calibration and evaluation purposes.

\paragraph{Coreset Selection.}

Benchmark subset selection can largely be viewed as an application of coreset selection, which often has been studied  in the context of curating training datasets. Typically, methods strive to select samples that are informative~\citep{el2n}, diverse~\citep{abbas2023semdedupdataefficientlearningwebscale, sener2018active, badge2019, bhatt-etal-2024-experimental}, aligned with a target~\citep{das2026matched, kothawade2021similar}, or some combination of these objectives~\citep{bukharin-zhao-2024-data, das2026matched, maharana2023d2pruningmessagepassing, badge2019}. Methods can utilize signals from the model as it is being trained as in active learning~\citep{craig, sener2018active, badge2019, el2n, kothawade2021similar, glister}, or can use signals from an extrinsic model (such as embeddings)~\citep{abbas2023semdedupdataefficientlearningwebscale, bhatt-etal-2024-experimental, bukharin-zhao-2024-data, das2026matched}. Submodularity has been used for data subset selection since~\cite{lin2009-submod-active-seq}.
Many of these methods can be applied directly to benchmark compression, particularly those that rely on extrinsic signals. In the setting of benchmark reduction, such signals can take the form of semantic embeddings or model evaluation scores. We find that semantic embeddings are generally more effective despite being significantly cheaper to acquire.

\section{Methodology}
\subsection{Facility Location (FL)}
\label{sec:method}
\paragraph{Problem Setup.}
The evaluation-unsupervised setting introduced above can be formalized as follows. Let $\mathcal{B}$ be the set of benchmarks, $V$ be the all benchmark samples (e.g., prompts plus metadata) and $\mathcal{C}$ the set of capability categories. Each item $i \in V$ belongs to exactly one benchmark $b(i) \in \mathcal{B}$, and each benchmark belongs to exactly one capability category $c \in \mathcal{C}$. We denote by $\mathcal{B}_c \subseteq \mathcal{B}$ the subset of benchmarks belonging to capability $c$, so that $\{\mathcal{B}_c\}_{c \in \mathcal{C}}$ partitions $\mathcal{B}$. Given a budget $k\ll n$, the selector must choose a coreset $A\subseteq V$ with $|A|=k$ without model-related information. Thus the selection may only depend on benchmark sample information, represented by an embedding $e_i$ for each $i$, and the induced nonnegative similarity score $w_{ij} \geq 0$ between any two items $i$ and $j$.

\paragraph{Objective.}
Given $w_{ij}$, a nonnegative similarity score between embeddings $e_i$ and $e_j$, the facility location function $f$ measures how well a subset $A$ represents $V$:
\begin{equation}
\label{eq:facility_location_function}
    f(A) = \sum_{i \in V} \max_{j \in A} \, w_{ij}
\end{equation}
In~\Eqref{eq:facility_location_function}, each element $i \in V$ (a ``client'') is represented by its most similar element $j \in A$ (its ``facility''), i.e., the $j \in A$ maximizing $w_{ij}$. In our task, we seek to select a subset that maximizes $f$:
\begin{equation}
\label{eq:facility_location_problem}
    A^* = \operatorname*{argmax}_{\substack{A \subseteq V \\ |A| = k}} \, f(A)
\end{equation}

Maximizing $f$ incentivizes selecting subsets $A$ whose elements collectively cover all of $V$ well. Since $f$ is a well-known submodular function, a simple greedy algorithm that iteratively adds the element with the largest marginal gain achieves a $(1 - 1/e)$ approximation guarantee~\citep{nemhauser1978analysis}. We provide a primer on submodular functions in the appendix. All experiments were performed using~\cite{bilmes2026-submarine}.

\paragraph{Semantic Embeddings.}
For each $i \in V$, we build a text representation by concatenating (i) the problem prompt, and (2) ground-truth and evaluation metadata fields. The second part is benchmark-specific: for instruction-following benchmarks it contains the executable instruction constraints; for knowledge benchmarks it contains the canonical answer text; for math benchmarks it contains the final answer; for coding benchmarks it contains oracle tests or solutions; and for agentic benchmarks it contains tool schema, task state, and expected actions. Appendix~\ref{app:experiment_details} lists the source fields used for each benchmark. We serialize structured fields instead of using plain text so that fine-grained format differences remain visible in the similarities. We embed each concatenated string with \textsc{Qwen3-Embedding-4B} to get $e_i=\mathrm{LLM}(\mathrm{text}_i)$. We left-pad and truncate to at most 8192 tokens.

\paragraph{Similarity Matrix Construction.}
Choosing the right similarity matrix is a critical design choice for
facility location to be effective. We define the similarity between samples
$i$ and $j$ as the (clipped) cosine similarity of their embeddings,
$w_{ij} = \max\!\left(0, \cos(e_i, e_j)\right)$. Cosine similarity matches the training objective of embedding models, and the clipping enforces $w_{ij} \geq 0$
which ensures that $f$ is submodular.

With a dense
similarity matrix, however, $f$ saturates quickly. Specifically, a single well-placed
facility attains nonzero similarity to most of $V$, so the first few
selections capture nearly all of the attainable objective value and the
marginal gains of subsequent elements collapse by several orders of
magnitude. Beyond saturation, the
dense matrix requires $O(n^2)$ memory, which can be substantially expensive for large
prompt sets. To ameliorate both issues, we sparsify the matrix, retaining
only the top-$t$ entries of each row, with $t = \lceil \alpha n / k \rceil$
and all other entries set to zero as is standard in~\cite{bilmes2026-submarine}. Sparsification also reduces
memory from $O(n^2)$ to $O(\alpha n^2 / k)$, allowing us to scale to much
larger datasets.

The parameter $\alpha$ interpolates between two failure
modes, both visible in Figure~\ref{fig:gain_curves}, which shows the
marginal gains of the greedy selection across sparsification levels. When
$\alpha$ is too small, each selected facility covers only a handful of
clients and these coverage sets rarely overlap, so every candidate yields a
nearly identical gain and the selection order
becomes largely arbitrary. When $\alpha$ is too large, the kernel
approaches density and saturation reappears, with gains collapsing within
the first few dozen selections. We therefore treat $\alpha$ as a
hyperparameter and tune it on a validation set, as described in
Section~\ref{sec:evaluation}.

\subsection{Evaluation}
\label{sec:evaluation}

We evaluate coreset fidelity by how faithfully the selected subset predicts
per-model benchmark scores relative to the full evaluation suite.

\paragraph{Prediction.}
Let $A_b = \{i \in A : b(i) = b\}$ be the selected items from benchmark $b$.
The predicted score of model $m$ on benchmark $b$ is
$\hat{s}^b_m(A) = |A_b|^{-1}\sum_{i \in A_b} y^m_i$,
where $y^m_i$ is the score of model $m$ on item $i$;
benchmarks with $A_b = \emptyset$ are excluded from all averages.
The overall predicted score aggregates across covered benchmarks:
\begin{equation}
\label{eq:pred}
    \hat{s}_m(A) \;=\; \frac{1}{|\mathcal{B}(A)|} \sum_{b \in \mathcal{B}(A)} \hat{s}^b_m(A),
\end{equation}
where $\mathcal{B}(A) = \{b \in \mathcal{B} : A_b \neq \emptyset\}$ is the set
of covered benchmarks.
The ground-truth $s_m$ is the same average computed over all items.
For capability $c$ (one of $|\mathcal{C}|{=}5$ categories), the predicted
capability score $\hat{s}^c_m(A)$ is the mean of $\hat{s}^b_m(A)$ over
benchmarks $b \in \mathcal{B}_c$ that are covered; the ground-truth
$s^c_m$ is defined analogously over all items.

\paragraph{Metrics.}
Our two primary metrics are \emph{mean capability relative error} (Cap-MRE)
and \emph{overall mean relative error} (MRE):
\begin{align}
\label{eq:cap_mre}
    \mathrm{Cap\text{-}MRE}(A) &\;=\; \frac{1}{|\mathcal{C}|} \sum_{c \in \mathcal{C}}
        \frac{1}{M} \sum_{m=1}^{M}
        \frac{\bigl|\hat{s}^c_m(A) - s^c_m\bigr|}{s^c_m}, \\
\label{eq:mre}
    \mathrm{MRE}(A) &\;=\; \frac{1}{M} \sum_{m=1}^{M}
        \frac{\bigl|\hat{s}_m(A) - s_m\bigr|}{s_m}.
\end{align}
Cap-MRE measures per-capability score fidelity averaged across all categories; MRE is the average of the overall benchmark fidelity. Since the overall score is itself an average over capabilities, the two metrics
aggregate the same per-capability errors very differently. MRE sums the signed
deviations before taking a magnitude, while Cap-MRE takes the magnitude within
each capability first. The overall error can be
smaller, so a subset that over-represents some capabilities and under-represents others lets these signed errors cancel, making MRE look near-perfect even when
every capability is badly estimated. Therefore, it is better to look at both Cap-MRE and MRE for deciding on the subset.

\section{Experiments}

\subsection{Experiment Setup}

To overcome the scale limitation of previous work, we newly sample  and evaluate models responses on over 61,498 prompts from 35 recently developed highly-valued benchmarks, covering five capability categories: agentic tool use, coding, instruction following, knowledge, and mathematics. We benchmark 18 frontier LLMs drawn from seven different model families or providers (Anthropic/Claude, DeepSeek, Google/Gemini, Moonshot/Kimi, OpenAI/GPT, Qwen, and xAI/Grok). We will release the large scale benchmark data for future research. The detailed benchmarks and models we use are presented in Appendix \ref{app:experiment_details}.

In total, we collect 1,106,964 entries in the (model, sample) matrix. Due to checkpoint unavailability and other force majeure factors, we end up having 47,521 missing entries (4.29\%). We impute missing entries with NeuMF~\citep{he2017neural}, an item-response theory (IRT) learning model, trained only on observed scores, with logit
\begin{equation}
\label{eq:neumf_logit}
\begin{split}
z_i^m = {}& b_0+b_i+c_m+\phi_i^\top\psi_m \\
          & +\,\mathrm{MLP}([\phi_i,\psi_m,\phi_i\odot\psi_m]),
\end{split}
\end{equation}
and binary cross-entropy training objective $\mathcal{L}_{\mathrm{NeuMF}}=\sum_{(i,m)\in\Omega}\mathrm{BCE}(y_i^m,\sigma(z_i^m))$. $\Omega$ is the set of observed model-sample pairs, $y_i^m$ is the observed score, $b_0$ is global bias, $b_i$ is sample bias, $c_m$ is model bias, and $\phi_i,\psi_m$ are the learned sample/model embeddings. The training process also yields sample/model embeddings that can be used for the IRT-based benchmark compression algorithm baseline \citep{polo2024tinybenchmarks}. We will refer them as IRT embedding in the following sections.

We use a 3-fold cross-validation over the $M{=}18$ models, partitioning them into three equal folds of six. In each round, two folds serve as the calibration split and the remaining fold as the held-out test split. Missing entries for the 12 models are replaced by NeuMF predictions. The models are used to tune the hyperparameters of the coreset selection algorithm, e.g., the $\alpha$ parameter for the similarity matrix that is used by FL. Held-out models do not have learned model embeddings, so any remaining missing entries for held-out models are filled with that model's observed mean only when constructing the fold's complete evaluation matrix. We use the average Cap-MRE on the calibration split for hyperparameter tuning, and report test split performance averaged over the three folds.

\subsection{Baselines}
We categorize the benchmark compression algorithm baselines into two families: score-based and diversity-based methods.


\paragraph{Score-Based Methods.}
These methods rank samples by a scalar score and select the top-$k$ with no coupling between selected items.

\begin{itemize}[leftmargin=*]
    \item \textbf{Hardness (IRT).} fits a one-parameter (1-PL) item-response theory model to the observed model-response matrix, obtaining a difficulty parameter $\beta_i$ per prompt~\citep{polo2024tinybenchmarks}, and selects the $k$ most difficult prompts. 
    \item \textbf{Point-biserial correlation (PBSC).} \citep{joshi2026datbench} computes the pointwise-biserial correlation $r_i$ between item scores $\{y^m_i\}_{m=1}^{M}$ and aggregate model scores $\{s_m\}_{m=1}^{M}$, and selects the $k$ prompts with the highest correlation. 
\end{itemize}
Formally, for every $A \subseteq V$ and $a \in A$ let $m_a \in \mathbb{R}$ be a score obtained using either of the approaches, then $f(A) = \sum_{a \in A} m_a$. 

\paragraph{Diversity-based Methods.}
These methods explicitly model interactions between candidate items to promote coverage.

\begin{itemize}[leftmargin=*]
    \item \textbf{Determinantal Point Processes (DPP).} A DPP assigns probability proportional to $\det(K_A)$ for every $A \subseteq V$, where $K$ is a positive-semidefinite kernel matrix~\citep{kulesza2012determinantal}. We use a greedy algorithm that iteratively adds the item with the highest marginal log-determinant gain, that is, MAP inference. Following our method, all DPP variants use sample text embeddings $e_i$. We consider four kernels: a \textbf{covariance kernel}~\citep{smola2026submodular} $K_{ij} = \tilde{e}_i^{\top}\tilde{e}_j$ with mean-centered embeddings $\tilde{e}_i = e_i - \bar{e}$ (\textbf{DPP-Cov}); a \textbf{cosine kernel} $K_{ij} = \cos(e_i, e_j)$ (\textbf{DPP-Cos}); and a \textbf{dot-product kernel} $K_{ij} = e_i^{\top} e_j$ (\textbf{DPP-Dot}). For either of the cases, the objective takes the form where $f(A) = \log(\det(K_A))$.  Inspired by \cite{JMLR:v9:krause08a, smola2026submodular, iyer2021submodular}, we also consider a mutual information-based variant, which is $\hat{f}(A) = I_f(A; V\setminus A)$, where $I_f(A;B) = f(A) + f(B) - f(A\cup B)$. For our DPP setup, this is $\hat{f}(A) = \log(\det K_{A})+\log(\det K_{V\setminus A}) - \log(\det K_{V})$ , which is an instance of non-monotone submodular function. For the Covariance, Cosine, and Dot Product kernels, this yields three additional baselines: \textbf{DPP-Cov-MI}, \textbf{DPP-Cos-MI}, and \textbf{DPP-Dot-MI}, with ``MI" referring to mutual information. 

    \item \textbf{$k$-Center.} The greedy $k$-Center algorithm iteratively selects the sample farthest from all currently selected items, minimizing the maximum distance from any sample to its nearest selected center~\citep{sener2018active}. \footnote{Please note that this objective is, in fact, a minmax facility location function.} To optimize the above NP-hard object, we follow the proposed greedy methods, which provide a 2-multiplicative approximation to the optimal selection~\citep{Cook2009-ci}. 

    \item \textbf{nn-$k$means Clustering.} We run the $k$-means clustering algorithm and assign each cluster centroid's nearest sample as its representative. For both $k$-Center and Clustering, to remain consistent with TinyBenchmarks~\citep{polo2024tinybenchmarks}, we evaluate two embedding variants: (i) \textbf{text embeddings} $e_i$ from \textsc{Qwen3-Embedding-4B}, identical to those used by facility location and DPP; and (ii) \textbf{IRT embeddings} $\phi_i$, the prompt latent vectors learned by our NeuMF model (\S\ref{sec:evaluation}), which encode the observed per-prompt response pattern across the training models.
\end{itemize}

\begin{figure}[t]
\centering
\includegraphics[width=\columnwidth]{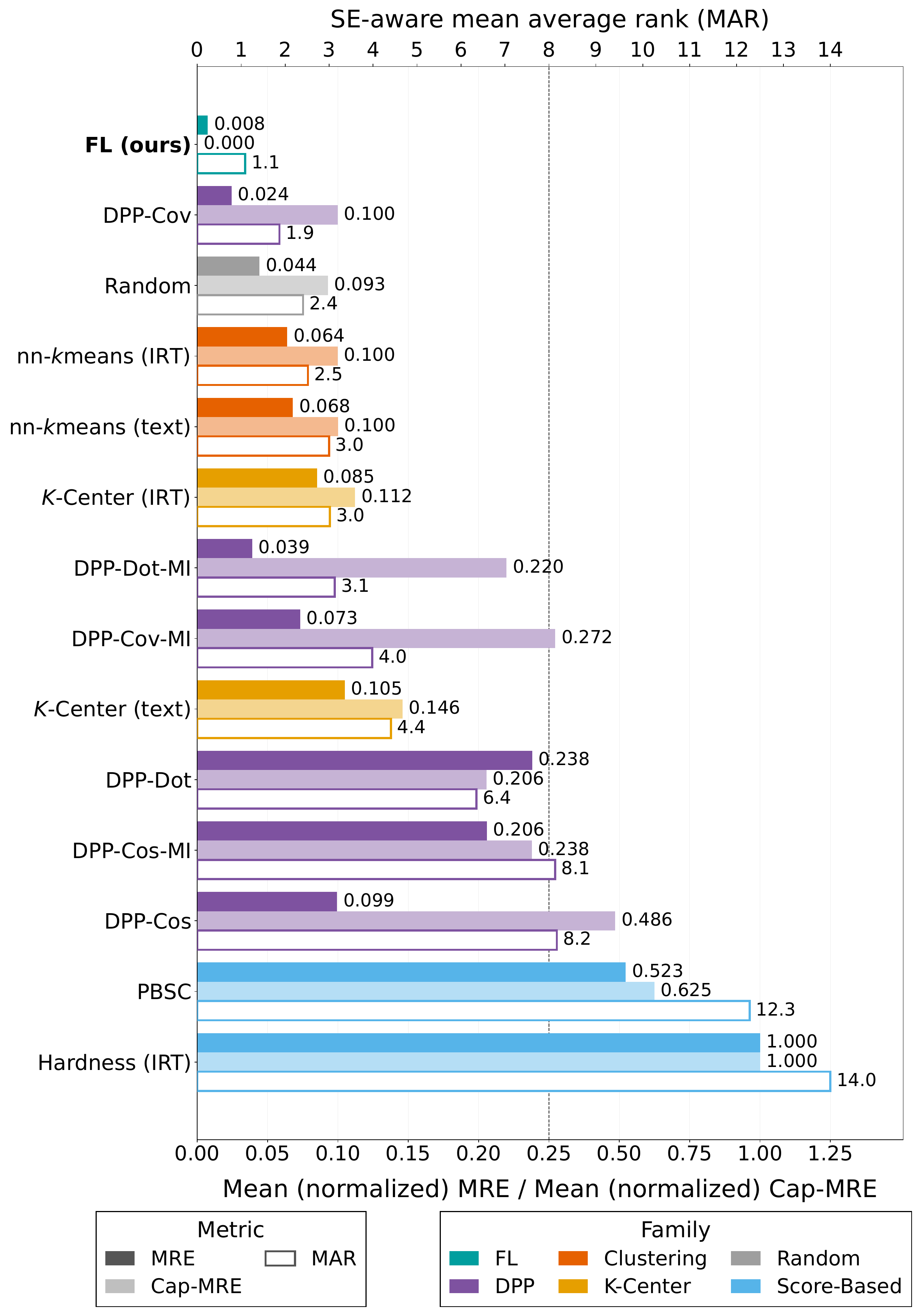}
\caption{The lower $x$-axis reports mean normalized MRE and mean normalized Cap-MRE, where scores are linearly normalized to $[0,1]$ within each budget and then averaged across all budgets $k \in [70, 200]$. The upper $x$-axis reports the SE-aware mean average rank jointly over MRE and Cap-MRE, where a method is ranked above another only when their error intervals do not overlap. \textbf{Lower values indicate better performance across all metrics.}. Methods are sorted top-to-bottom by MAR. The dashed vertical line marks the change in the bottom $x$-axis scale used to include the high-error baselines without compressing the remaining methods. Colors indicate method family.}
\label{fig:norm_combined}
\end{figure}

\subsection{Main Results}

\paragraph{FL almost always performs best across all budgets.}
Tables~\ref{tab:combined_results} report MRE and Cap-MRE (mean $\pm$ SE across folds) at four representative budgets: $k \in \{80, 120, 160, 200\}$. Full per-budget plots across the entire range in increments of 5, $k \in [70, 200]$, are provided in Appendix~\ref{app:full_budget} (Figures~\ref{fig:full_mre} and~\ref{fig:full_capmre}). Moreover, Figure~\ref{fig:norm_combined} summarize performance across all coreset budgets up to $k=200$ with the mean normalized score for each method on MRE and Cap-MRE, respectively. We generally see a trend that FL almost always performs best across all budgets. FL achieves the lowest normalized MRE (0.03) and a near-zero normalized Cap-MRE, beating all baselines even those using model response information.

Figure~\ref{fig:norm_combined} also reports the SE-aware mean average rank (MAR) jointly over MRE and Cap-MRE for all methods.
FL achieves a MAR of 1.1, close to the theoretical minimum of 1, indicating near-uniform dominance across budgets and metrics.
DPP-Cov is the strongest competitor with a MAR of 1.8, followed by Clustering at 2.4. We also observe that other variants of DPP underperform, which could be attributed to their tendency to rely heavily on a single benchmark rather than selection across the benchmark.  Score-based baselines occupy the bottom two positions (MAR 11.3 and 13.0). This is unsurprising: without an explicit diversity objective, they tend to concentrate selections within a single benchmark, failing to achieve the broad coverage needed for low relative error.

\begin{table*}[t]
\centering
\small
\tabcolsep=0.055cm
\renewcommand{\arraystretch}{1.2}
\definecolor{flbluebg}{RGB}{235,243,255}

\begin{tabular*}{\textwidth}{@{}l@{\extracolsep{\fill}}*{8}{r}@{}}
\toprule[1.5pt]
\multirow{2}{*}{\textbf{Method}} 
& \multicolumn{2}{c}{$k=80$}
& \multicolumn{2}{c}{$k=120$}
& \multicolumn{2}{c}{$k=160$}
& \multicolumn{2}{c}{$k=200$} \\
\cmidrule(lr){2-3} \cmidrule(lr){4-5} \cmidrule(lr){6-7} \cmidrule(lr){8-9}
& \multicolumn{1}{c}{\makebox[0pt][c]{\hspace{6pt}Cap-MRE $\downarrow$}}
& \multicolumn{1}{c}{MRE $\downarrow$}
& \multicolumn{1}{c}{\makebox[0pt][c]{\hspace{6pt}Cap-MRE $\downarrow$}}
& \multicolumn{1}{c}{MRE $\downarrow$}
& \multicolumn{1}{c}{\makebox[0pt][c]{\hspace{6pt}Cap-MRE $\downarrow$}}
& \multicolumn{1}{c}{MRE $\downarrow$}
& \multicolumn{1}{c}{\makebox[0pt][c]{\hspace{6pt}Cap-MRE $\downarrow$}}
& \multicolumn{1}{c}{MRE $\downarrow$} \\
\midrule[0.75pt]

\rowcolor{flbluebg}
\quad Random
& $0.213\err{.022}$ & $0.084\err{.005}$
& $0.179\err{.021}$ & $0.091\err{.009}$
& $0.175\err{.020}$ & $0.083\err{.005}$
& $0.174\err{.020}$ & $0.079\err{.004}$ \\
\midrule[0.75pt]

\multicolumn{9}{l}{\textit{Score-Based} \cite{yao2025measure, joshi2026datbench}} \\
\quad PBSC
& $0.486\err{.028}$ & $0.400\err{.026}$
& $0.456\err{.033}$ & $0.379\err{.045}$
& $0.480\err{.029}$ & $0.375\err{.054}$
& $0.469\err{.030}$ & $0.371\err{.044}$ \\

\quad Hardness (IRT)
& $0.692\err{.029}$ & $0.693\err{.042}$
& $0.683\err{.024}$ & $0.683\err{.020}$
& $0.681\err{.022}$ & $0.672\err{.040}$
& $0.684\err{.018}$ & $0.678\err{.032}$ \\
\midrule[0.75pt]

\multicolumn{9}{l}{\textit{$k$-Center} \cite{sener2018active, polo2024tinybenchmarks}} \\
\quad $k$-Center (IRT)
& $0.213\err{.026}$ & $0.112\err{.035}$
& $0.198\err{.030}$ & $0.124\err{.027}$
& $0.188\err{.029}$ & $0.091\err{.019}$
& $0.162\err{.021}$ & $0.089\err{.022}$ \\

\rowcolor{flbluebg}
\quad $k$-Center (text)
& $0.290\err{.038}$ & $0.139\err{.021}$
& $0.241\err{.044}$ & $0.142\err{.019}$
& $0.183\err{.037}$ & $0.123\err{.005}$
& $0.150\err{.023}$ & $0.106\err{.001}$ \\
\midrule[0.75pt]

\multicolumn{9}{l}{\textit{nn-$k$means clustering} \cite{polo2024tinybenchmarks}} \\
\quad nn-$k$means (IRT)
& $\underline{0.199\err{.034}}$ & $0.118\err{.028}$
& $0.186\err{.033}$ & $0.076\err{.020}$
& $0.203\err{.037}$ & $0.126\err{.011}$
& $0.196\err{.024}$ & $0.121\err{.022}$ \\

\rowcolor{flbluebg}
\quad nn-$k$means (text)
& $0.254\err{.054}$ & $0.080\err{.007}$
& $0.217\err{.076}$ & $0.089\err{.004}$
& $0.180\err{.035}$ & $0.124\err{.006}$
& $\underline{0.141\err{.032}}$ & $0.061\err{.012}$ \\
\midrule[0.75pt]

\multicolumn{9}{l}{\textit{DPP} \cite{kulesza2012determinantal, JMLR:v9:krause08a, smola2026submodular}} \\
\rowcolor{flbluebg}
\quad DPP-Cov
& $0.266\err{.046}$ & $0.077\err{.017}$
& $\underline{0.175\err{.013}}$ & $\underline{0.066\err{.004}}$
& $\underline{0.151\err{.004}}$ & $\underline{0.062\err{.009}}$
& $0.159\err{.009}$ & $0.075\err{.001}$ \\

\rowcolor{flbluebg}
\quad DPP-Cov-MI
& $0.313\err{.081}$ & $0.114\err{.018}$
& $0.326\err{.088}$ & $0.117\err{.023}$
& $0.246\err{.076}$ & $0.081\err{.007}$
& $0.232\err{.077}$ & $0.080\err{.005}$ \\

\rowcolor{flbluebg}
\quad DPP-Cos
& $0.405\err{.059}$ & $0.129\err{.020}$
& $0.395\err{.076}$ & $0.110\err{.016}$
& $0.381\err{.053}$ & $0.118\err{.013}$
& $0.395\err{.062}$ & $0.118\err{.013}$ \\

\rowcolor{flbluebg}
\quad DPP-Cos-MI
& $0.260\err{.043}$ & $0.202\err{.020}$
& $0.254\err{.044}$ & $0.176\err{.021}$
& $0.279\err{.045}$ & $0.175\err{.021}$
& $0.256\err{.037}$ & $0.184\err{.026}$ \\

\rowcolor{flbluebg}
\quad DPP-Dot
& $0.277\err{.103}$ & $0.238\err{.006}$
& $0.264\err{.106}$ & $0.229\err{.013}$
& $0.224\err{.087}$ & $0.178\err{.013}$
& $0.227\err{.088}$ & $0.186\err{.012}$ \\

\rowcolor{flbluebg}
\quad DPP-Dot-MI
& $0.244\err{.034}$ & $\underline{0.060\err{.015}}$
& $0.317\err{.053}$ & $0.085\err{.003}$
& $0.227\err{.063}$ & $0.105\err{.017}$
& $0.202\err{.068}$ & $\mathbf{0.048\err{.006}}$ \\
\midrule[0.75pt]

\rowcolor{flbluebg}
\quad FL (ours)
& $\mathbf{0.141\err{.030}}$ & $\mathbf{0.050\err{.015}}$
& $\mathbf{0.139\err{.018}}$ & $\mathbf{0.059\err{.013}}$
& $\mathbf{0.127\err{.017}}$ & $\mathbf{0.055\err{.009}}$
& $\mathbf{0.113\err{.019}}$ & $\underline{0.053\err{.001}}$ \\
\bottomrule[1.5pt]
\end{tabular*}

\caption{
Cap-MRE and MRE at key budgets $k \in \{80, 120, 160, 200\}$; \textbf{the lower the better}.
\textbf{Bold} indicates the best result, and \underline{underline} indicates the second-best. \colorbox{flbluebg}{Blue background} represents evaluation-unsupervised algorithms. FL almost always performs best
across all budgets. 
}
\label{tab:combined_results}
\end{table*}

\paragraph{Possible Pitfalls of IRT based embeddings.}
The method proposed in tinyBenchmarks~\citep{polo2024tinybenchmarks} learns prompt embeddings
from IRT models: each prompt is assigned a latent
difficulty parameter estimated from how models respond to it, and these
parameters are used to cluster prompts into representative groups.
The quality of these embeddings, therefore, depends critically on having enough
models to reliably estimate the latent item parameters.
In our setting, we have only $M{=}18$ models shared across all 35 benchmarks making the IRT
parameter estimates possibly noisy. Thus the potential instability of IRT embeddings trained on small models set may explain their failure.

To demonstrate this concretely, we measure how much the IRT-induced similarity
structure changes across the three model folds.
For each benchmark, we compute the Spearman correlation $\rho$ between the
pairwise IRT similarity matrices produced by each pair of folds (three pairs
total) and report the mean and standard error across pairs.
As shown in Table~\ref{tab:irt_stability}, $\rho$ ranges from $0.30$
(Nexus Bench) to $0.92$ (Follow Bench), with
many benchmarks below $0.7$. Semantic embeddings are computed from sample text alone and are therefore
identical regardless of which models are available, making them more reliable for facility location. This indicates IRT's high dependency on large amount of model responses, which is infeasible to collect in reality.

\begin{figure}[t]
\centering
\includegraphics[width=\columnwidth]{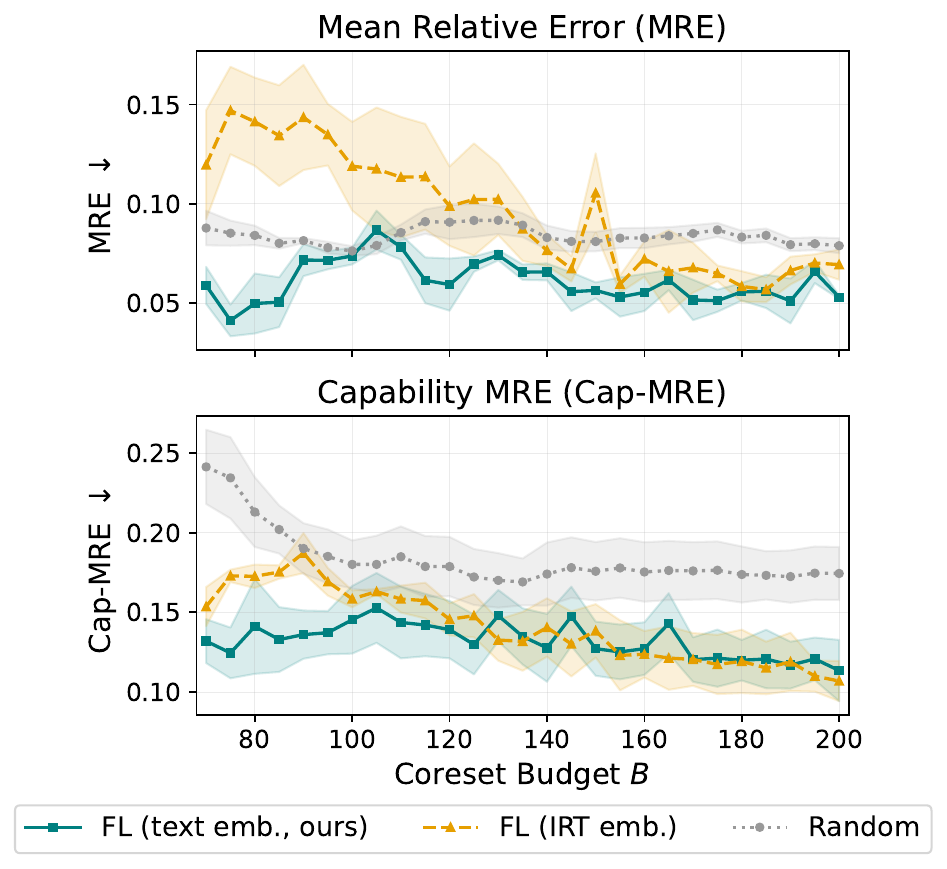}
\caption{FL with text embeddings vs.\ FL with IRT embeddings vs.\ random, across all budgets $k \in [70, 200]$; \textbf{the lower the better}. Shaded bands show standard error across folds.}
\label{fig:fl_irt_ablation}
\end{figure}

\paragraph{Text vs.\ IRT embeddings on facility location.}
Figure~\ref{fig:fl_irt_ablation} directly compares FL using text embeddings against FL using IRT embeddings, alongside random selection as a reference.
On MRE, FL (text) outperforms FL (IRT) substantially across all budgets, with the gap most pronounced at small budgets ($k \leq 120$): at $k{=}80$, FL (text) achieves $0.050 \pm 0.015$ versus $0.141 \pm 0.022$ for FL (IRT).
Crucially, the SE bands for FL (IRT) on MRE are consistently wider, reflecting the instability in IRT-derived embeddings: because latent item parameters vary across folds, FL selects different subsets each time, producing high variance in test error.
On Cap-MRE the two methods converge at larger budgets -- FL (IRT) reaches $0.107 \pm 0.013$ at $k{=}200$ versus $0.113 \pm 0.019$ for FL (text), but the SE bands overlap substantially throughout, and FL (IRT) requires access to model response data, making it inapplicable in the evaluation-unsupervised setting.
Taken together, text embeddings provide a more stable and practically accessible basis for FL without sacrificing capability-level coverage.

\begin{figure}[t]
\centering
\includegraphics[width=\columnwidth]{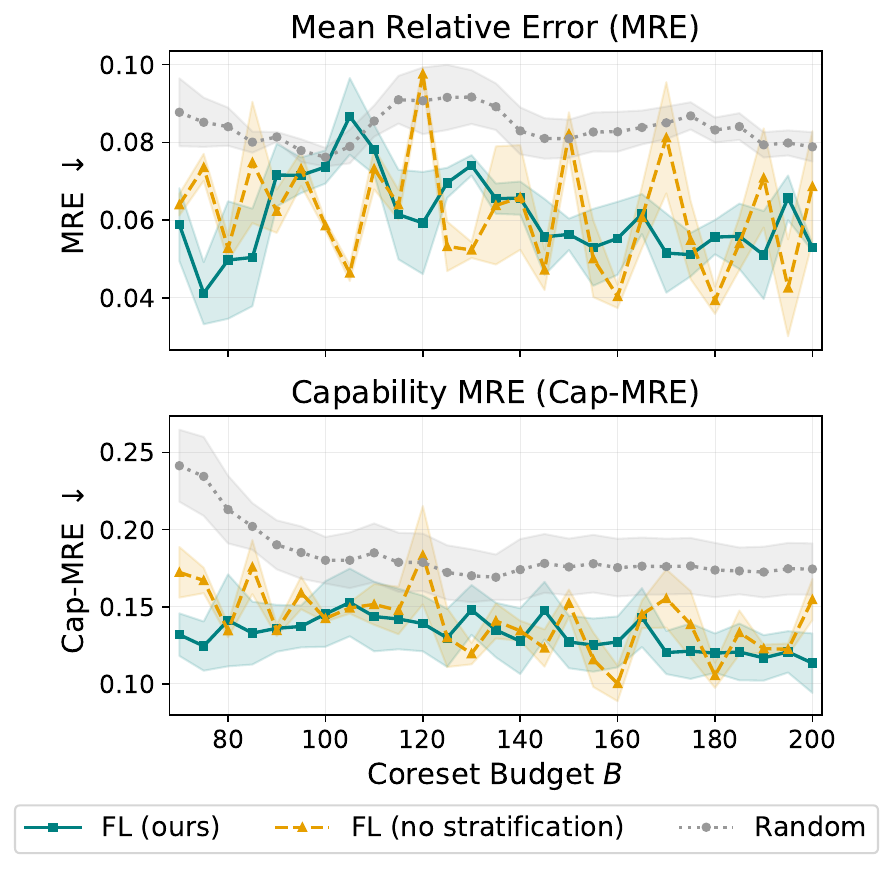}
\caption{FL with block-diagonal similarity (ours) vs.\ FL without benchmark stratification (no stratification) vs.\ random, across all budgets $k \in [70, 200]$; \textbf{the lower the better}. Shaded bands show standard error across folds. Both the FL variants outperform random.}
\label{fig:fl_stratification_ablation}
\end{figure}

\paragraph{Effect of block-diagonal similarity structure.}
In our FL setup, the pairwise similarity matrix is constrained to be block-diagonal according to benchmark identity: prompts from different benchmarks have zero similarity, preventing the greedy marginal-gain computation from trading coverage across benchmarks.
This benchmark stratification is a deliberate design choice to prevent the coreset from concentrating on a single benchmark.
That said, it is not a necessary condition for facility location to work: a well-tuned FL objective can still select a diverse set of examples even under severe cross-benchmark size imbalance.
Figure~\ref{fig:fl_stratification_ablation} compares FL (ours, with stratification) against FL without stratification, and confirms that removing the constraint does not catastrophically degrade performance — both variants comfortably outperform random selection across the full budget range.
As future work, one could leverage the unstratified FL solution to quantify benchmark similarity: the overlap in facility coverage between two benchmarks, when the similarity matrix spans all prompts jointly, provides a natural proxy for the mutual information between two benchmarks.

\subsection{Benchmark Level Subset Selection}
A separate line of work studies submodular selection not over samples within a
benchmark, but over entire \emph{benchmarks} within a leaderboard~\citep{smola2026submodular}.
\paragraph{Setting.} 
Given a score matrix $\mathcal{S}$ of $M$ models $\times$ $N$ benchmarks, the goal is to select a 
small set $A$ of $k$ benchmarks, such that an unseen model's scores
on the unselected complement $\bar A = V \setminus A$ can be reconstructed from its
scores on $A$ alone. That work estimates a benchmark-level correlation matrix
$\hat\Sigma$ and greedily selects $A$ to maximize either differential entropy or
mutual information with the complement, which is precisely the case where one uses a
DPP with the covariance matrix as a kernel.

Partitioning the estimated parameters along $A$ and $\bar A$,
\begin{equation}
\hat\mu = \begin{bmatrix}\hat\mu_A\\[2pt]\hat\mu_{\bar A}\end{bmatrix},
\qquad
\hat\Sigma =
\begin{bmatrix}
\hat\Sigma_{AA} & \hat\Sigma_{A\bar A}\\[2pt]
\hat\Sigma_{\bar A A} & \hat\Sigma_{\bar A\bar A}
\end{bmatrix},
\end{equation}
a test model $i$ is evaluated only on the selected benchmarks, and its unselected
scores are imputed by the Gaussian conditional mean
\begin{equation}
\hat B_{i\bar A}
= \hat\mu_{\bar A}
+ \hat\Sigma_{\bar A A}\,
  \hat\Sigma_{AA}^{-1}
  \bigl(B_{iA} - \hat\mu_A\bigr),
\end{equation}
We evaluate each method with
two complementary metrics: test $R^2$, which measures how well an unseen model's
held-out scores are recovered, and the residual variance fraction, which measures
how much benchmark variance remains unexplained after conditioning on $A$.

We report test $R^2$ in standardized space, so the constant predictor
$\hat B_{i\bar A}=\hat\mu_{\bar A}$ attains $R^2 = 0$:
\begin{equation}
R^2 = 1 -
\frac{\sum_{i}\sum_{j\in\bar A}\bigl(\hat B_{ij}-B_{ij}\bigr)^2}
     {\sum_{i}\sum_{j\in\bar A}\bigl(B_{ij}-\bar x_j\bigr)^2},
\end{equation}
where $\bar x_j$ is the training-set mean of benchmark $j$. A high $R^2$ at small
$k$ implies that the $k$ selected benchmarks suffice to recover the full leaderboard
profile of an unseen model. The residual variance fraction, in contrast, is a
model-independent quantity derived directly from the conditional covariance of
$\bar A$ given $A$, i.e.\ the Schur complement
\begin{equation}
\hat\Sigma_{\bar A\mid A}
= \hat\Sigma_{\bar A\bar A}
- \hat\Sigma_{\bar A A}\,
  \hat\Sigma_{AA}^{-1}
  \hat\Sigma_{A\bar A} .
\end{equation}

\begin{equation}
\mathrm{RV}(A)
= \frac{\operatorname{tr}\!\bigl(\hat\Sigma_{\bar A\mid A}\bigr)}
       {\operatorname{tr}\!\bigl(\hat\Sigma\bigr)} .
\end{equation}
Smaller values indicate that the selected benchmarks explain more of the total
leaderboard variance, and $1-\mathrm{RV}(A)$ is the corresponding explained-variance
fraction.

\paragraph{Experimental Setup.} We evaluate on two leaderboards, MMLU ($5451$ models, $57$ sub-benchmarks) and MTEB
($263$ models, $56$ sub-benchmarks), with a $70/30$ train/test split. The correlation matrix $\hat\Sigma$ is estimated on the training
split and used for both selection and imputation across all methods. While indeed one can do the text embedding-based approach as done in Section~\ref{sec:method}, we keep setup consistent with \citet{smola2026submodular} for fair comparison. Moreover, in this particular setup, since the number of models is many-fold compared to Section~\ref{sec:method}, the covariance matrix is reliable. We run FL on
the same $\hat\Sigma$, applying the truncation
$\hat\Sigma^{+}_{ij} = \max(\hat\Sigma_{ij}, 0)$ since FL requires non-negative pairwise similarities, unlike DPP, which requires positive semidefiniteness; imputation still uses the original $\hat\Sigma$, so the truncation affects only which benchmarks are selected, never how scores are predicted.

\begin{figure}[t]
\centering
\includegraphics[width=\columnwidth]{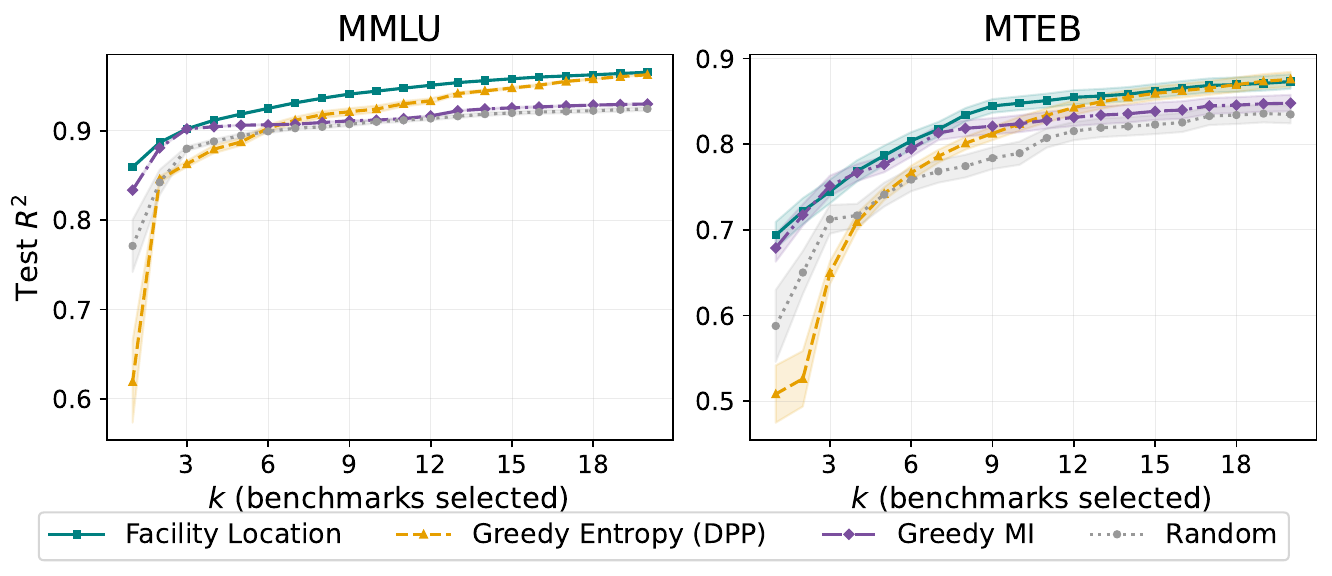}
\caption{Benchmark-level selection on MMLU and MTEB (same protocol as~\citep{smola2026submodular}), with FL run on the identical score-derived similarity matrix $\Sigma$ used by greedy entropy and greedy MI. FL matches or beats both DPP-style objectives almost everywhere, and dominates throughout on MTEB according to mean test $R^2$.}
\label{fig:smola_comparison}

\vspace{\floatsep}
\includegraphics[width=\columnwidth]{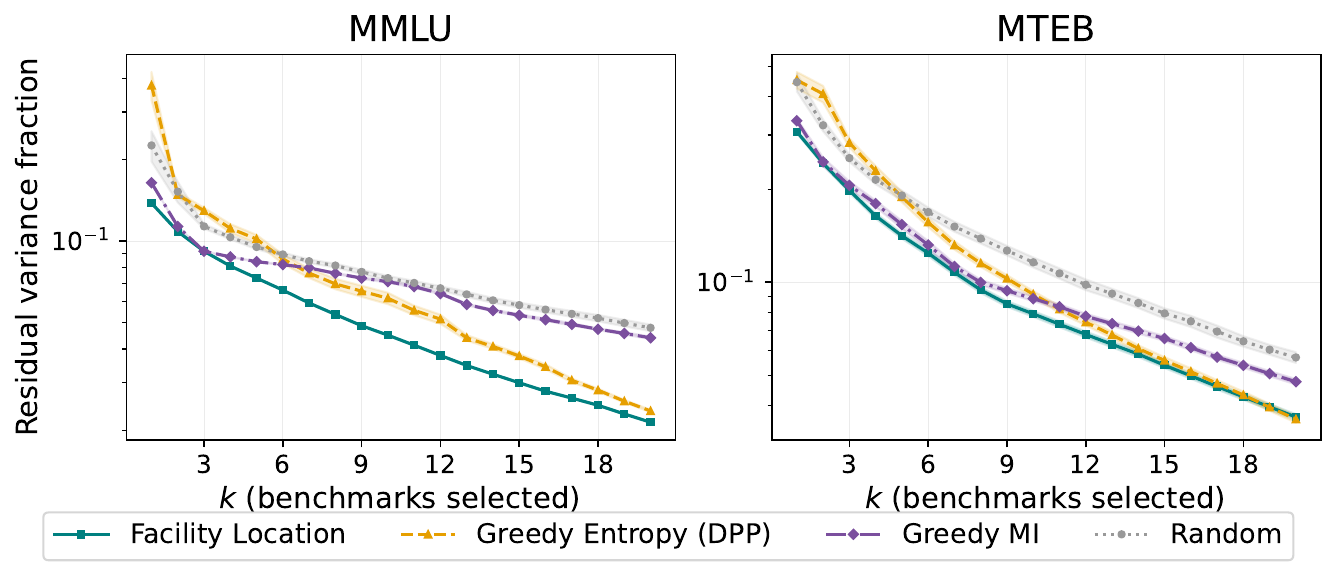}
\caption{Benchmark-level selection on MMLU and MTEB (same protocol
as~\citep{smola2026submodular}), with FL run on the identical score-derived
similarity matrix $\Sigma$ used by greedy entropy and greedy MI. FL achieves the lowest residual variance fraction across the entire range on both leaderboards according to mean residual
variance fraction.}
\label{fig:smola_residvar}
\end{figure}
\paragraph{Results.}Figure~\ref{fig:smola_comparison} reports test $R^2$ as a function of the number of
selected benchmarks $k$ on MMLU and MTEB. FL is the strongest method across nearly the entire range, and its advantage is largest in the small-$k$ regime that
matters most for cheap evaluation: on MMLU, it reaches $R^2\approx0.86$ with a single
benchmark and $\approx0.92$ by $k=4$, where Greedy Entropy (DPP-Cov) still trails badly ($R^2\approx0.61$ at $k=1$) since the first element can be arbitrarily bad. It recovers only for larger $k$, converging to FL by
$k\approx18$--$20$; greedy MI (DPP-Cov-MI) is competitive at small $k$, particularly on MTEB, but plateaus below FL and Greedy Entropy ($R^2\approx0.83$--$0.85$) at larger $k$. Overall, a small number of FL-selected benchmarks suffices to recover the full leaderboard profile of an unseen model:
$k=4$ benchmarks out of $57$ already yield $R^2>0.9$ on MMLU. Figure~\ref{fig:smola_residvar} shows scores of residual variance fraction, where FL again achieves the best across the entire range on both leaderboards. In appendix~\ref{sec: app-extended_smola}, we expand DPPs on the choice of kernel to Gaussian kernels, and again observe that FL achieves the best across the entire range on both leaderboards. 

\begin{figure}[t]
\centering
\includegraphics[width=\columnwidth]{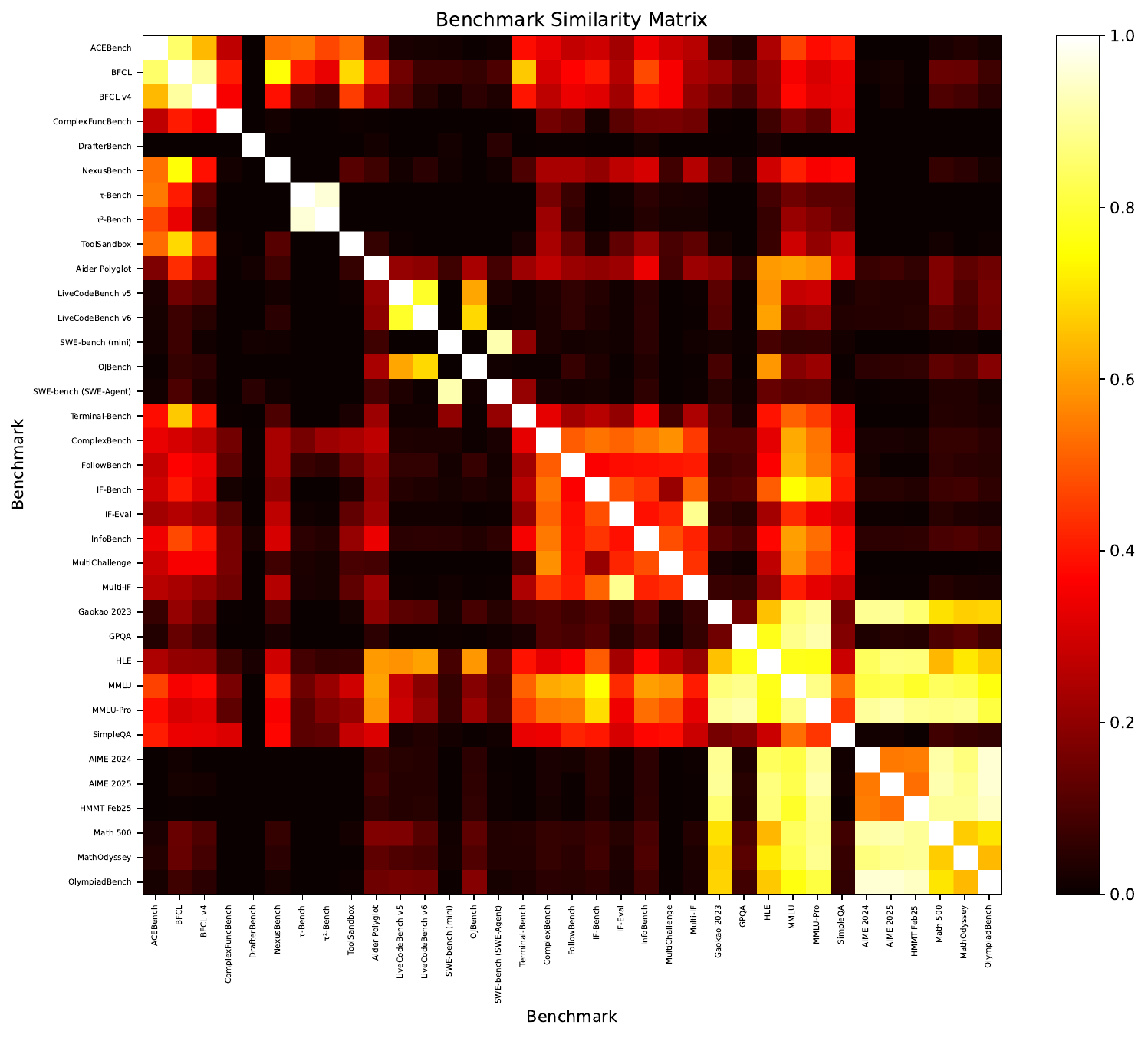}
\caption{Normalized benchmark--benchmark similarity matrix $M$ (Eq.~\ref{eq:bench_smi}),
computed purely from prompt embeddings. A bright entry
$M_{bb'}$ indicates that benchmarks $b$ and $b'$ redundantly cover the same region of
embedding space; this matrix is the similarity used by the benchmark-level
facility-location greedy that produces the ordering in Table~\ref{tab:bench_order}.}
\label{fig:bench_similarity_matrix}
\end{figure}

\paragraph{Benchmark level similarity using Submodular Mutual Information.} The benchmark-level selection of \citet{smola2026submodular} operates in the
\emph{evaluation-supervised} regime: it estimates a benchmark--benchmark similarity
matrix from the covariance ${\Sigma}$ of a dense model--benchmark score matrix.
This raises a natural question: when such scores are unavailable, how can one
construct the similarity matrix? Here we show that the \emph{same} facility-location based approach can build a benchmark-level similarity matrix and order the benchmarks by the greedy FL ordering, but in a fully
\emph{unsupervised} manner that depends on \emph{prompt text}. The key idea
is to replace the score covariance with the submodular mutual information (SMI) of the prompt-level FL function. Let $A_b = \{i \in V : b(i) = b\}$ be the set of prompts belonging to benchmark
$b\in\mathcal{B}$, and let $f$ be the prompt-level facility-location function induced
by the cosine similarity $w_{ij}=\max(0,\cos(e_i,e_j))$ over the prompt embeddings.
For any two prompt subsets $A$ and $B$, the submodular mutual information
$I_f(A;B) = f(A) + f(B) - f(A\cup B)$ measures how strongly the two sets overlap in
the regions of embedding space they cover. Since
$I_f(A_b;A_{b'}) \le \min\!\big(f(A_b),f(A_{b'})\big)$, we normalize it to obtain a
benchmark--benchmark similarity matrix $M$,
\begin{equation}
M_{bb'} \;=\; \frac{I_f(A_b;A_{b'})}{\min\!\big(f(A_b),\,f(A_{b'})\big)},
\label{eq:bench_smi}
\end{equation}
so that a large $M_{bb'}$ indicates that benchmarks $b$ and $b'$ redundantly cover
the same region of embedding space. Figure~\ref{fig:bench_similarity_matrix} visualizes
$M$ over the full set of benchmarks. 

We then define a second, benchmark-level
facility-location function $G$ over the ground set $\mathcal{B}$, using $M$ as its
similarity matrix,
\begin{equation}
G(S) \;=\; \sum_{b\in\mathcal{B}} \max_{b'\in S} M_{bb'}, \qquad S \subseteq \mathcal{B},
\label{eq:bench_fl}
\end{equation}
and order the benchmarks using the greedy order based on the largest conditional gain. This produces a \emph{conditional} ordering: the first benchmark is the most
representative of the entire suite, and each subsequent benchmark is the most
representative one \emph{given} those already selected. Table~\ref{tab:bench_order} (in appendix~\ref{app:benchmark_smi}) reports the full ordering. The greedy order first selects \textsc{MMLU}, the broadest and
largest knowledge benchmark, and the remaining order tracks the semantic structure of
the suite: near-duplicate benchmarks are deferred to the end once a
representative has been chosen. \textsc{MMLU-Pro} (the second-largest benchmark, with
$12{,}032$ prompts) drops to rank $25$ because \textsc{MMLU} already covers it;
likewise \textsc{BFCL v4} (rank $29$) follows \textsc{BFCL} (rank $2$),
\textsc{LiveCodeBench v5} (rank $22$) follows \textsc{v6} (rank $3$),
\textsc{AIME 2024} (rank $34$) follows \textsc{AIME 2025} (rank $8$), and
$\tau$-\textsc{Bench} (rank $35$) follows $\tau^2$-\textsc{Bench} (rank $5$). This provides an additional application of the facility location function through the lens of submodular mutual information.

\section{Conclusion}
We studied \emph{evaluation-unsupervised benchmark coreset selection}, where a coreset is selected using no previous model evaluation outcomes and information only from the prompts themselves, in contrast to prior evaluation-supervised methods that rely on model responses or evaluation scores. Our main finding is that facility location
using semantic embeddings is a strong and reliable choice in this regime. It attains the
lowest MRE and Cap-MRE across the full
budget range $k \in [70, 200]$ on a new large-scale suite of 35 heterogeneous
benchmarks, 18 frontier LLMs, and 61K prompts. Facility location outperforms twelve
baselines and random selection, with the covariance-kernel DPP its closest competitor
and difficulty-based selectors, biserial correlation the weakest. This indicates that
diversity, rather than difficulty, is what matters for high fidelity.
Further analysis reveals that IRT embeddings learned from a sparse model score matrix can be very noisy, whereas text embeddings provide a model-independent basis.
Beyond the evaluation-unsupervised setting, we also evaluate facility location under the evaluation-supervised setting for benchmark subset selection, where abundant model scores are available;
there too, facility location outperforms submodular baselines including DPP and its mutual-information variant. Lastly, we provide a case where one can construct a benchmark--benchmark similarity matrix using submodular mutual information (see~\ref{app:benchmark_smi}) with a facility location function, motivating future work for evaluation-unsupervised benchmark subset selection. Overall, these results suggest that facility location is an effective and practical objective for benchmark coreset construction.

\FloatBarrier

\bibliography{iclr2025_conference}
\newpage
\appendix

\section{A Primer on Submodular Functions}
\label{app:submodular_primer}

Submodular functions are set functions that exhibit
diminishing returns. Formally, let $V$ be the ground set of elements and let $f: 2^V \to \mathbb{R}$ be a set function.
For two subsets $A \subseteq B \subseteq V$ and an element $v \in V \setminus B$,
$f$ is \emph{submodular} if
\begin{equation}
\label{eq:submodular}
    f(A \cup \{v\}) - f(A) \;\geq\; f(B \cup \{v\}) - f(B).
\end{equation}
The quantities $f(A \cup \{v\}) - f(A)$ and $f(B \cup \{v\}) - f(B)$ are the
\emph{marginal gains} of adding $v$ to $A$ and to $B$, respectively. Condition~\eqref{eq:submodular}
states that the marginal gain of an element to a smaller set is at least its
marginal gain to a larger superset. In contrast to score-based functions, which
assign an independent value to each element and evaluate a subset by summing those
values (so that $f(A) = \sum_{v \in A} m_v$ for per-element scores $m_v$),
submodular functions capture interactions among the selected elements and are
therefore well suited to modeling notions of diversity and coverage. A set function is additionally \emph{monotone} if $f(A) \leq f(B)$
whenever $A \subseteq B$; the facility location function of
\Eqref{eq:facility_location_function} is both monotone and submodular. For
benchmark coreset selection we maximize such a function subject to the cardinality
budget $k$,
\begin{equation}
\label{eq:sm_opt}
    A^* \;=\; \operatorname*{argmax}_{\substack{A \subseteq V \\ |A| = k}} \, f(A),
\end{equation}
recovering the problem in \Eqref{eq:facility_location_problem}. Although
\Eqref{eq:sm_opt} is NP-hard in general, monotone and normalized submodular function maximization under a
cardinality constraint admits a simple greedy algorithm that starts from
$A = \emptyset$ and repeatedly adds the element of largest marginal gain,
$\operatorname*{argmax}_{v \in V \setminus A} f(v \mid A)$, attaining a
$(1 - 1/e)$ approximation of the optimum~\citep{nemhauser1978analysis};
accelerated variants such as lazy and stochastic greedy preserve comparable
guarantees while scaling to much larger ground
sets~\citep{mirzasoleiman2015lazier}. See~\citet{bilmes2022submodularity} for more
details.

\section{Experiment Details}
\label{app:experiment_details}

\paragraph{Benchmarks.}
We evaluate 35 benchmarks grouped into five categories:
\begin{itemize}[leftmargin=*]
    \item \textbf{Agentic Tool Use}: ACEBench~\citep{chen2025acebench}, BFCL~\citep{patilberkeley}, ComplexFuncBench~\citep{zhong2025complexfuncbench}, DrafterBench~\citep{li2025drafterbench}, MultiChallenge~\citep{sirdeshmukh2025multichallenge}, NexusBench~\citep{nexusbench}, $\tau$-Bench~\citep{yao2024tau}, $\tau^2$-Bench~\citep{barres2025tau}, ToolSandbox~\citep{lu2024toolsandbox}.
    \item \textbf{Coding}: LiveCodeBench v5/v6~\citep{jain2024livecodebench}, OJBench~\citep{wang2025ojbench}, Terminal-Bench~\citep{tbench_2025}, SWE-bench~\citep{jimenez2023swe}.
    \item \textbf{Instruction following}: ComplexBench~\citep{wen2024benchmarking}, FollowBench~\citep{jiang2023followbench}, IF-Bench~\citep{pyatkin2025generalizing}, IF-Eval~\citep{zhou2023instruction}, InfoBench~\citep{qin2024infobench}, MultiChallenge~\citep{sirdeshmukh2025multichallenge}, Multi-IF~\citep{he2024multi}.
    \item \textbf{Knowledge}: MMLU~\citep{hendrycks2020measuring}, MMLU-Pro~\citep{wang2024mmlu}, GPQA~\citep{rein2024gpqa}, SimpleQA~\citep{wei2024measuring}, HLE~\citep{phan2025humanity}, Gaokao 2023~\citep{zhang2023evaluating}.
    \item \textbf{Math}:AIME 2024/2025~\citep{aops_aime_2025}, HMMT Feb25~\citep{balunovic2025matharena}, Math 500~\citep{hendrycks2021measuring}, MathOdyssey~\citep{fang2025mathodyssey}, OlympiadBench~\citep{he2024olympiadbench}.
\end{itemize}

\paragraph{Models.}
We evaluate on 18 models: \textsc{DeepSeek-R1-0528}~\citep{guo2025deepseek}, \textsc{DeepSeek-V3-0324}~\citep{liu2024deepseek}, \textsc{DeepSeek-V3.1}~\citep{liu2024deepseek}, \textsc{DeepSeek-V3.1 thinking}~\citep{liu2024deepseek}, \textsc{Kimi-K2-Instruct}~\citep{team2025kimi}, \textsc{Qwen3-235B-A22B-FP8}~\citep{yang2025qwen3}, \textsc{Qwen3-235B-A22B-Instruct-2507-FP8}~\citep{yang2025qwen3}, \textsc{Claude-4-Sonnet}~\citep{anthropic2025claude4}, \textsc{Claude-4-Sonnet thinking}~\citep{anthropic2025claude4}, \textsc{Gemini-2.5-Flash thinking}~\citep{comanici2025gemini25pushingfrontier}, \textsc{Gemini-2.5-Pro thinking}~\citep{comanici2025gemini25pushingfrontier}, \textsc{GPT-4.1}~\citep{hurst2024gpt}, \textsc{GPT-4o-20240806}~\citep{hurst2024gpt}, \textsc{GPT-4o-mini}~\citep{hurst2024gpt}, \textsc{GPT-5}~\citep{singh2026openaigpt5card}, \textsc{Grok-4}~\citep{xai2025grok4}, \textsc{O3-high}~\citep{openai2025o3o4mini}, and \textsc{O4-mini-high}~\citep{openai2025o3o4mini}.

\paragraph{Embedding fields.}
Each concat embedding uses the prompt plus the benchmark target fields below. ``Prompt'' refers to the original prompt field name used to construct the prompt shown to the model. ``Target'' refers to the additional ground-truth or evaluation-target field name appended to the prompt before embedding. When the target field is listed as none, the embedding is prompt-only for that benchmark.

\begin{itemize}[leftmargin=*]
    \item \textbf{ACEBench}: Prompt: \texttt{question}, \texttt{function}, \texttt{initial\_config}. Target: \texttt{rubric.ground\_truth}.
    \item \textbf{BFCL}: Prompt: \texttt{question} with tool/function definitions from \texttt{function}, \texttt{functions}, or \texttt{tools}, and task-state fields \texttt{initial\_config}. Target: \texttt{ground\_truth}.
    \item \textbf{ComplexFuncBench}: Prompt: user turns in \texttt{conversations[*].content}. Target: none.
    \item \textbf{DrafterBench}: Prompt: \texttt{Instruction}. Target: \texttt{Groundtruth}.
    \item \textbf{MultiChallenge}: Prompt: \texttt{CONVERSATION}. Target: \texttt{TARGET\_QUESTION}.
    \item \textbf{NexusBench}: Prompt: task text from \texttt{Input}, \texttt{prompt}, \texttt{user\_query}, \texttt{generated\_question}, and tool-schema fields from \texttt{json\_tools} or \texttt{fc\_definition} when present. Target: \texttt{Output}, \texttt{outputs.reference}, \texttt{ground\_truth}, \texttt{answer}, \texttt{fncall}, or  \texttt{modified\_correct\_ground\_truth}.
    \item \textbf{$\tau$-Bench}: Prompt: \texttt{instruction}. Target: \texttt{actions}.
    \item \textbf{$\tau^2$-Bench}: Prompt: \texttt{user\_scenario.instructions}, \texttt{initial\_state}. Target: \texttt{evaluation\_criteria}.
    \item \textbf{ToolSandbox}: Prompt: starting message \texttt{Scenario.starting\_context}, setting-state fields \texttt{cellular}, \texttt{wifi}, \texttt{location\_service}, \texttt{low\_battery\_mode}, \texttt{latitude}, \texttt{longitude}, plus \texttt{tool\_allow\_list} and \texttt{tool\_augmentation\_list}. Target: none.
    \item \textbf{Aider Polyglot}: Prompt: \texttt{instruction.md}. Target: none.
    \item \textbf{LiveCodeBench v5/v6}: Prompt: v5 \texttt{prompt}; v6 \texttt{question\_content} and \texttt{starter\_code}. Target: v5 \texttt{verification\_info.ground\_truth}; v6 \texttt{public\_test\_cases}.
    \item \textbf{OJBench}: Prompt: \texttt{prompt}. Target: none.
    \item \textbf{Terminal-Bench}: Prompt: \texttt{instruction.md}. Target: \texttt{solution/solve.sh}.
    \item \textbf{SWE-bench}: Prompt: \texttt{problem\_statement}. Target: \texttt{patch}.
    \item \textbf{ComplexBench}: Prompt: \texttt{instruction\_en}. Target: \texttt{scoring\_questions[*].question\_en} or \texttt{scoring\_questions[*].question}.
    \item \textbf{FollowBench}: Prompt: \texttt{instruction}. Target: \texttt{target}.
    \item \textbf{IF-Bench}: Prompt: \texttt{prompt}. Target: \texttt{instruction\_id\_list}, \texttt{kwargs}.
    \item \textbf{IF-Eval}: Prompt: \texttt{prompt}. Target: \texttt{instruction\_id\_list}, \texttt{kwargs}.
    \item \textbf{InfoBench}: Prompt: \texttt{instruction}, \texttt{input}. Target: \texttt{decomposed\_questions}.
    \item \textbf{Multi-IF}: Prompt: \texttt{turn\_1\_prompt}, \texttt{turn\_2\_prompt}, \texttt{turn\_3\_prompt}. Target: \texttt{turn\_1\_instruction\_id\_list}, \texttt{turn\_1\_kwargs}, \texttt{turn\_2\_instruction\_id\_list}, \texttt{turn\_2\_kwargs}, \texttt{turn\_3\_instruction\_id\_list}, \texttt{turn\_3\_kwargs}.
    \item \textbf{MMLU}: Prompt: \texttt{question}, \texttt{choices}. Target: \texttt{choices[answer]}.
    \item \textbf{MMLU-Pro}: Prompt: \texttt{question}, \texttt{options}. Target: \texttt{options[answer\_index]} or \texttt{answer}.
    \item \textbf{GPQA}: Prompt: \texttt{Question}. Target: \texttt{Correct Choice}.
    \item \textbf{SimpleQA}: Prompt: \texttt{problem}. Target: \texttt{answer}.
    \item \textbf{HLE}: Prompt: \texttt{question}. Target: \texttt{answer}.
    \item \textbf{Gaokao 2023}: Prompt: \texttt{question}. Target: \texttt{answer}.
    \item \textbf{AIME 2024/2025}: Prompt: \texttt{problem}. Target: AIME 2024 \texttt{reference\_solution} boxed answer or \texttt{expected\_answer}; AIME 2025 \texttt{expected\_answer}.
    \item \textbf{HMMT Feb25}: Prompt: \texttt{problem}. Target: \texttt{expected\_answer}.
    \item \textbf{Math 500}: Prompt: \texttt{problem}. Target: \texttt{answer}.
    \item \textbf{MathOdyssey}: Prompt: \texttt{problem\_statement}. Target: \texttt{expected\_answer}.
    \item \textbf{OlympiadBench}: Prompt: \texttt{question}. Target: \texttt{answer}.
\end{itemize}

\section{Analysis (Cont')}
\subsection{Full Per-Budget Results}
\label{app:full_budget}
\begin{figure*}[t]
\centering
\includegraphics[width=1.0\linewidth]{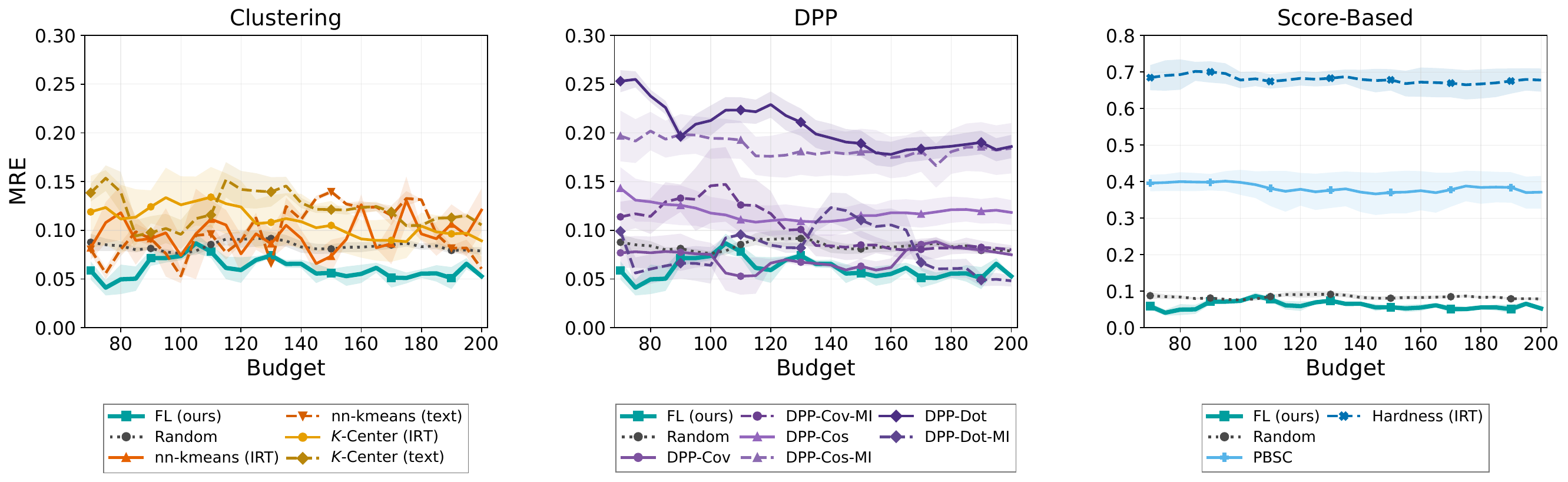}
\caption{Per-budget overall relative error (MRE) for all methods across budgets
$k \in [70, 200]$ in increments of $5$ (lower is better). Curves show the mean over
the three model folds; shaded bands denote $\pm$SE. Score-based baselines are
omitted (see text). FL (ours) attains the lowest MRE across essentially the entire
range, with DPP-Cov the closest competitor.}
\label{fig:full_mre}
\end{figure*}

\begin{figure*}[t]
\centering
\includegraphics[width=1.0\linewidth]{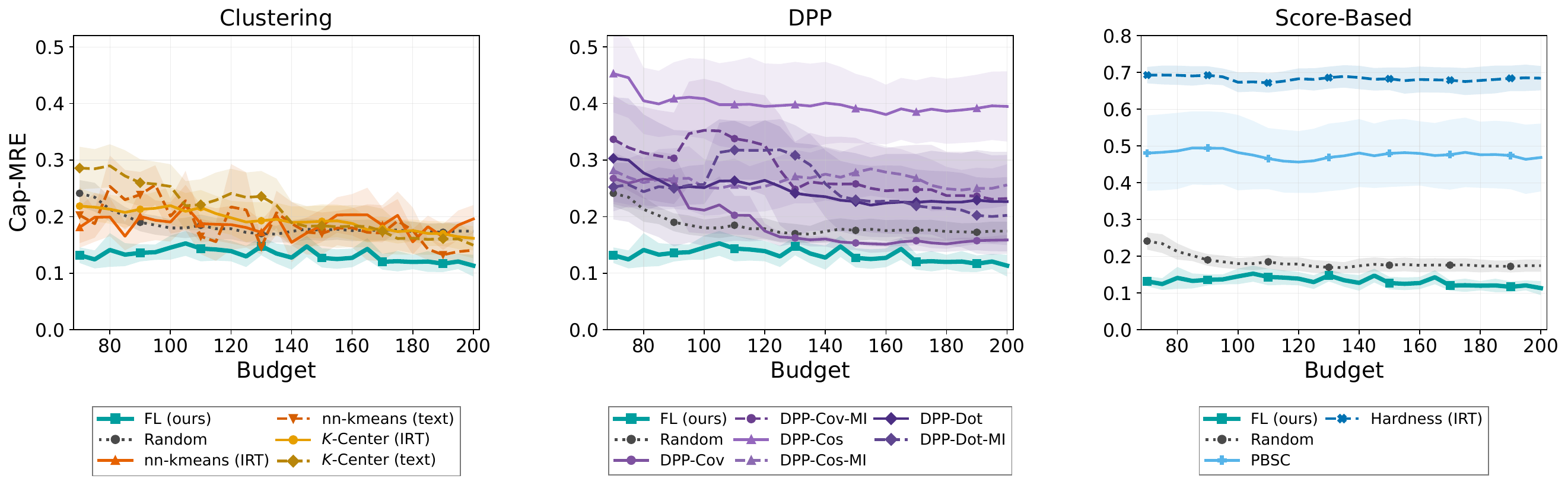}
\caption{Per-budget capability-wise relative error (Cap-MRE) for all methods across
budgets $k \in [70, 200]$ in increments of $5$ (lower is better). Same protocol as
Figure~\ref{fig:full_mre}. FL (ours) attains the lowest Cap-MRE across essentially
the entire range, while the cosine-kernel DPP variants are weakest.}
\label{fig:full_capmre}
\end{figure*}

Tables~\ref{tab:combined_results} report results at four
representative budgets, and Figure~\ref{fig:norm_combined} summarize performance aggregated across budgets. For completeness, we provide the
full per-budget curves here: Figure~\ref{fig:full_mre} reports overall relative
error (MRE) and Figure~\ref{fig:full_capmre} reports capability-wise relative
error (Cap-MRE), each evaluated at every budget $k \in [70, 200]$ in increments of
$5$. Every curve is the mean over the three model folds and the shaded bands denote
$\pm$SE. All twelve baselines are shown; the two score-based baselines (PBSC and Hardness) are omitted because
their errors are an order of magnitude larger and would compress the visible range
(they are included in Tables~\ref{tab:combined_results} and~\ref{tab:combined_results}).

\begin{figure}[t]
\centering
\includegraphics[width=\columnwidth]{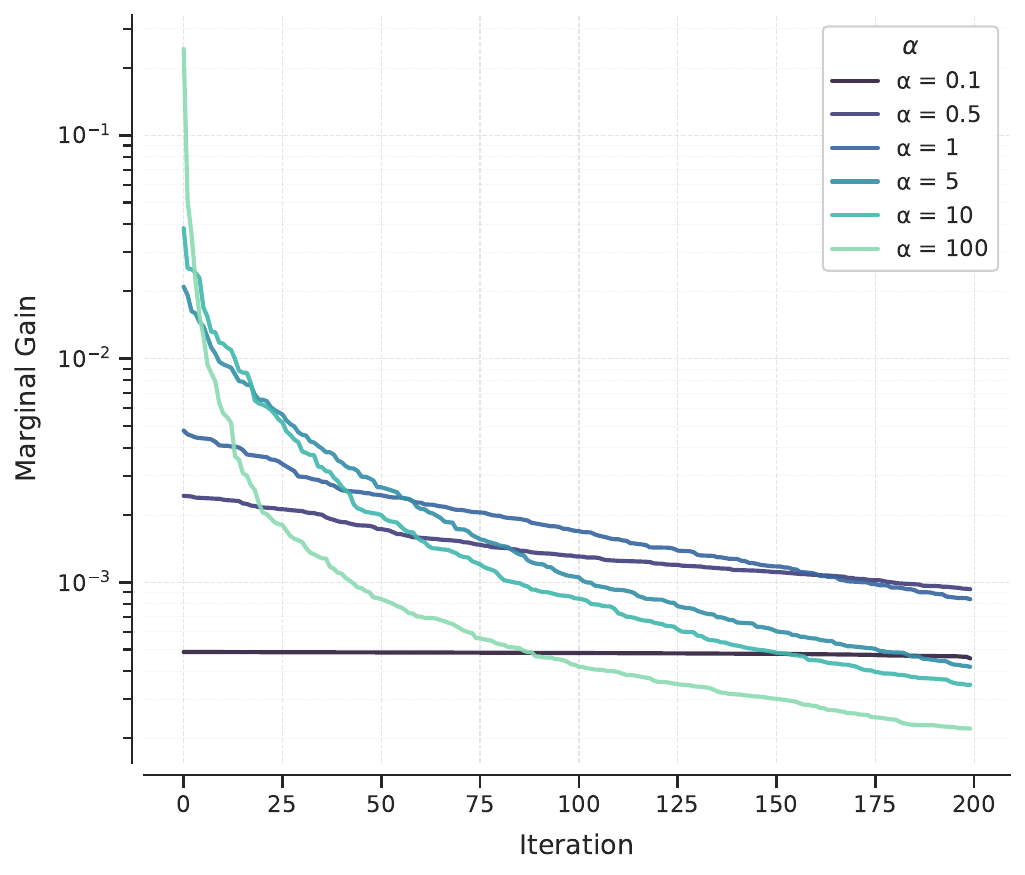}
\caption{Marginal gains of the greedy selection across sparsification levels $\alpha$. When $\alpha$ is too small, candidate gains are nearly identical and the selection order becomes arbitrary; when $\alpha$ is too large, the kernel approaches density and gains collapse within the first few dozen selections.}
\label{fig:gain_curves}
\end{figure}

\subsection{Analysis on IRT induced similarity}
We measure how much the IRT-induced similarity
structure changes across the three model folds.
For each benchmark, we compute the Spearman correlation $\rho$ between the
pairwise IRT similarity matrices produced by each pair of folds (three pairs
total) and report the mean and standard error across pairs.
As shown in Table~\ref{tab:irt_stability}, $\rho$ ranges from $0.30$
(Nexus Bench) to $0.92$ (Follow Bench), with
many benchmarks below $0.7$. Semantic embeddings are computed from sample text alone and are therefore
identical regardless of which models are available, making them more reliable for facility location. This indicates IRT's high dependency on large amount of model responses, which is infeasible to collect in reality.

\begin{table}[t]
\centering
\small
\setlength{\tabcolsep}{5pt}
\begin{tabular}{lrr}
\toprule[1.5pt]
\textbf{Benchmark} & $n$ & $\rho$ \\
\midrule[0.75pt]
NexusBench        & $1124$  & $0.298\err{0.059}$ \\
ToolSandbox        & $1032$  & $0.333\err{0.056}$ \\
AIME 2024             & $30$    & $0.367\err{0.128}$ \\
DrafterBench       & $1920$  & $0.433\err{0.044}$ \\
Aider Polyglot    & $225$   & $0.520\err{0.037}$ \\
AIME 2025              & $30$    & $0.530\err{0.031}$ \\
ComplexFuncBench & $1000$  & $0.557\err{0.026}$ \\
MultiChallenge      & $273$   & $0.608\err{0.025}$ \\
ComplexFuncBench        & $1150$  & $0.612\err{0.042}$ \\
$\tau$-Bench         & $165$   & $0.652\err{0.037}$ \\
HMMT Feb25         & $30$    & $0.657\err{0.050}$ \\
Multi-IF          & $4445$  & $0.673\err{0.033}$ \\
Math 500            & $500$   & $0.681\err{0.090}$ \\
MathOdyssey       & $387$   & $0.707\err{0.043}$ \\
SWE-bench (SWE-Agent)         & $500$   & $0.716\err{0.046}$ \\
$\tau^2$-Bench         & $278$   & $0.727\err{0.020}$ \\
IF-Eval            & $541$   & $0.735\err{0.027}$ \\
OlympiadBench       & $675$   & $0.753\err{0.048}$ \\
MMLU           & $14042$ & $0.772\err{0.040}$ \\
SWE-bench (mini-SWE-Agent)    & $500$   & $0.773\err{0.015}$ \\
GPQA           & $198$   & $0.797\err{0.011}$ \\
HLE           & $2158$  & $0.815\err{0.011}$ \\
InfoBench           & $500$   & $0.820\err{0.018}$ \\
OJBench          & $232$   & $0.831\err{0.026}$ \\
LiveCodeBench v5 & $166$   & $0.836\err{0.015}$ \\
Terminal-Bench    & $80$    & $0.840\err{0.036}$ \\
IF-Bench             & $294$   & $0.842\err{0.019}$ \\
SimpleQA       & $4326$  & $0.846\err{0.015}$ \\
BFCL v3               & $4441$  & $0.875\err{0.026}$ \\
MMLU-Pro       & $12032$ & $0.875\err{0.035}$ \\
ACEBench          & $1023$  & $0.876\err{0.004}$ \\
Gaokao 2023    & $385$   & $0.881\err{0.019}$ \\
LiveCodeBench v6 & $131$   & $0.899\err{0.017}$ \\
FollowBench         & $820$   & $0.923\err{0.002}$ \\
BFCL v4           & $5865$  & $0.923\err{0.011}$ \\
\bottomrule[1.5pt]
\end{tabular}
\caption{Per-benchmark stability of the IRT-induced similarity structure across
the three model folds. For each benchmark, we report the number of prompts $n$
and the mean Spearman correlation $\rho$ over three-fold pairs between the
pairwise IRT similarity matrices. Benchmarks are sorted by $\rho$; many fall
well below $0.6$, indicating that the performance-matrix-derived similarity
structure may be unstable when only $M{=}18$ models are available.}
\label{tab:irt_stability}
\end{table}

\subsection{Computational Cost Comparison} Beyond performance, FL and DPP are not equally cheap to scale. FL is much more efficient to compute, because of one doesn't even need a full $N\times N$ matrix, and can work with sparse, non-negative pairwise affinity. DPP, on the other hand, needs a symmetric, positive-semidefinite matrix for probabilities to make sense. While in this current setup, the ground set over which subset selection is performed is relatively small, the computational complexity of DPP-based methods can become prohibitive as the number of benchmarks or prompts increases.

\subsection{Extended Results with Gaussian Kernel}
\label{sec: app-extended_smola}
\begin{figure*}[t]
\centering
\includegraphics[width=\linewidth]{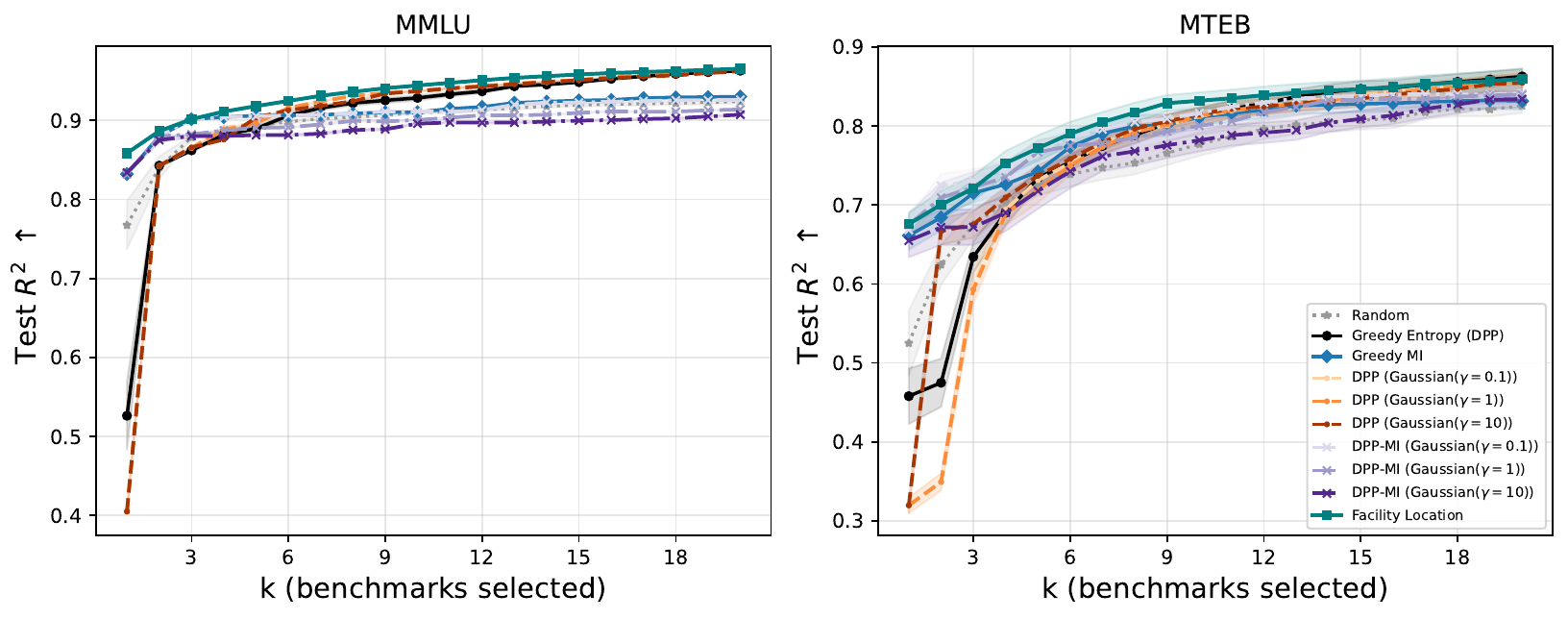}
\caption{Benchmark-level selection on MMLU and MTEB (same protocol as~\citep{smola2026submodular}), with FL run on the identical score-derived similarity matrix $\Sigma$ used by greedy entropy and greedy MI. FL matches or beats both DPP-style objectives almost everywhere, and dominates throughout on MTEB according to mean test $R^2$.}
\label{fig:smola_comparison_gaussian}
\end{figure*}

\begin{figure*}[t]
\centering
\includegraphics[width=\linewidth]{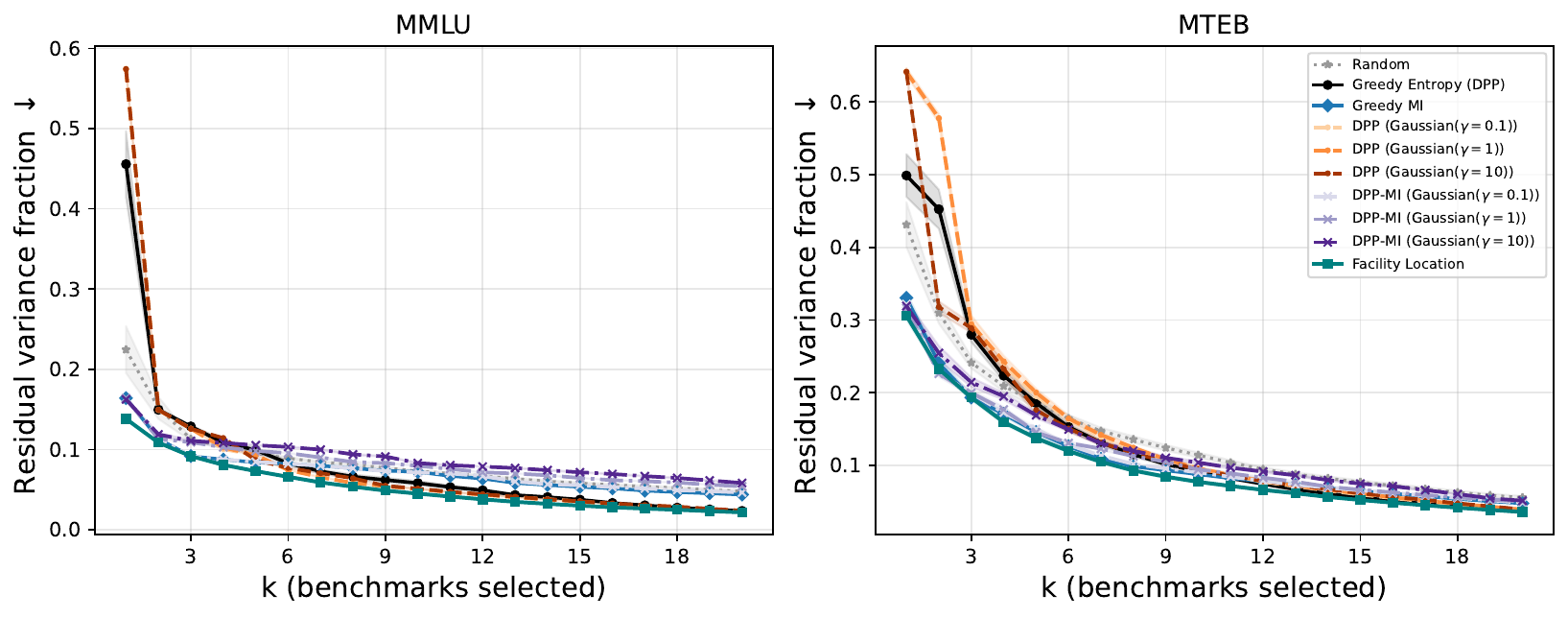}
\caption{Benchmark-level selection on MMLU and MTEB (same protocol
as~\citep{smola2026submodular}), with FL run on the identical score-derived
similarity matrix $\Sigma$ used by greedy entropy and greedy MI. FL achieves the lowest residual variance fraction across the entire range on both leaderboards according to mean residual
variance fraction.}
\label{fig:smola_resivar_gaussian}
\end{figure*}

While \citet{smola2026submodular} assumed benchmark performances as a joint Gaussian with covariance $\Sigma$, one can go beyond this assumption by assuming an explicit kernel form. More precisely, we model the similarity between two benchmarks $i, j$ as $K_{i, j} = \exp(- \gamma \|\mathcal{S}_{:, i}-\mathcal{S}_{:, j}\|^2)$, and compare it against facility location and existing DPP-based results here; we tune $\gamma \in \{0.1, 1, 10\}$. The evalutaion results are displayed in Figure~\ref{fig:smola_comparison_gaussian} and Figure~\ref{fig:smola_resivar_gaussian}. Our findings remain consistent with facility location function outperforming all the baselines judged using test-$R^2$ and residual variance fraction. 

\begin{table*}[!t]
\centering
\scriptsize
\setlength{\tabcolsep}{3pt}
\begin{tabular}{@{}l ccccc ccccc@{}}
\toprule
 & \multicolumn{5}{c}{\textbf{MMLU}} & \multicolumn{5}{c}{\textbf{MTEB}} \\
\cmidrule(lr){2-6}\cmidrule(lr){7-11}
Method & $k{=}1$ & $k{=}5$ & $k{=}10$ & $k{=}15$ & $k{=}20$ & $k{=}1$ & $k{=}5$ & $k{=}10$ & $k{=}15$ & $k{=}20$ \\
\midrule
Random & .767$_{\pm.031}$ & .893$_{\pm.003}$ & .910$_{\pm.002}$ & .919$_{\pm.002}$ & .924$_{\pm.002}$ & .525$_{\pm.041}$ & .727$_{\pm.015}$ & .777$_{\pm.015}$ & .807$_{\pm.012}$ & .826$_{\pm.009}$ \\
\midrule
Greedy Entropy (DPP) & .526$_{\pm.043}$ & .889$_{\pm.003}$ & .928$_{\pm.003}$ & .948$_{\pm.001}$ & .963$_{\pm.000}$ & .458$_{\pm.035}$ & .734$_{\pm.011}$ & .811$_{\pm.012}$ & \textbf{.847$_{\pm.010}$} & \textbf{.863$_{\pm.010}$} \\
Greedy MI & .832$_{\pm.002}$ & \underline{.906$_{\pm.001}$} & .910$_{\pm.001}$ & .925$_{\pm.001}$ & .930$_{\pm.001}$ & .661$_{\pm.018}$ & .742$_{\pm.016}$ & .812$_{\pm.011}$ & .828$_{\pm.011}$ & .832$_{\pm.011}$ \\
\addlinespace
DPP (Gaussian($\gamma{=}0.1$)) & .405$_{\pm.002}$ & .897$_{\pm.001}$ & \underline{.937$_{\pm.001}$} & \underline{.951$_{\pm.000}$} & \underline{.964$_{\pm.000}$} & .320$_{\pm.010}$ & .731$_{\pm.012}$ & .812$_{\pm.009}$ & .839$_{\pm.009}$ & \underline{.860$_{\pm.010}$} \\
DPP (Gaussian($\gamma{=}1$)) & .405$_{\pm.002}$ & .897$_{\pm.001}$ & .937$_{\pm.001}$ & .951$_{\pm.000}$ & .964$_{\pm.000}$ & .320$_{\pm.010}$ & .719$_{\pm.016}$ & .813$_{\pm.009}$ & .837$_{\pm.010}$ & .858$_{\pm.010}$ \\
DPP (Gaussian($\gamma{=}10$)) & .405$_{\pm.002}$ & .904$_{\pm.001}$ & .937$_{\pm.001}$ & .951$_{\pm.000}$ & .962$_{\pm.000}$ & .320$_{\pm.010}$ & .737$_{\pm.013}$ & .815$_{\pm.009}$ & .835$_{\pm.010}$ & .855$_{\pm.011}$ \\
\addlinespace
DPP-MI (Gaussian($\gamma{=}0.1$)) & .832$_{\pm.002}$ & .905$_{\pm.001}$ & .911$_{\pm.000}$ & .922$_{\pm.001}$ & .926$_{\pm.001}$ & .668$_{\pm.016}$ & \underline{.770$_{\pm.009}$} & \underline{.816$_{\pm.010}$} & .838$_{\pm.010}$ & .841$_{\pm.010}$ \\
DPP-MI (Gaussian($\gamma{=}1$)) & .832$_{\pm.002}$ & .891$_{\pm.001}$ & .900$_{\pm.001}$ & .910$_{\pm.001}$ & .914$_{\pm.001}$ & \underline{.675$_{\pm.015}$} & .767$_{\pm.015}$ & .800$_{\pm.011}$ & .833$_{\pm.011}$ & .840$_{\pm.011}$ \\
DPP-MI (Gaussian($\gamma{=}10$)) & \underline{.834$_{\pm.002}$} & .881$_{\pm.001}$ & .896$_{\pm.001}$ & .900$_{\pm.001}$ & .907$_{\pm.001}$ & .655$_{\pm.021}$ & .718$_{\pm.023}$ & .782$_{\pm.014}$ & .809$_{\pm.011}$ & .834$_{\pm.011}$ \\
\midrule
FL (ref) & \textbf{.859$_{\pm.001}$} & \textbf{.918$_{\pm.001}$} & \textbf{.944$_{\pm.000}$} & \textbf{.958$_{\pm.000}$} & \textbf{.965$_{\pm.000}$} & \textbf{.676$_{\pm.014}$} & \textbf{.772$_{\pm.016}$} & \textbf{.832$_{\pm.011}$} & \textbf{.847$_{\pm.011}$} & \underline{.860$_{\pm.012}$} \\
\bottomrule
\end{tabular}
\caption{Test $R^2$ ($\uparrow$) across budgets $k$ (20 splits). Best per column \textbf{bold}, second \underline{underlined}. facility location outperforms all the other baselines, even when the DPP variants are provided with a kernel with larger capacity.}
\label{tab:compare_test_r2}
\end{table*}

\begin{table*}[!t]
\centering
\scriptsize
\setlength{\tabcolsep}{3pt}
\begin{tabular}{@{}l ccccc ccccc@{}}
\toprule
 & \multicolumn{5}{c}{\textbf{MMLU}} & \multicolumn{5}{c}{\textbf{MTEB}} \\
\cmidrule(lr){2-6}\cmidrule(lr){7-11}
Method & $k{=}1$ & $k{=}5$ & $k{=}10$ & $k{=}15$ & $k{=}20$ & $k{=}1$ & $k{=}5$ & $k{=}10$ & $k{=}15$ & $k{=}20$ \\
\midrule
Random & .224$_{\pm.029}$ & .095$_{\pm.002}$ & .073$_{\pm.002}$ & .058$_{\pm.001}$ & .048$_{\pm.001}$ & .431$_{\pm.031}$ & .186$_{\pm.007}$ & .114$_{\pm.004}$ & .077$_{\pm.003}$ & .056$_{\pm.002}$ \\
\midrule
Greedy Entropy (DPP) & .456$_{\pm.041}$ & .099$_{\pm.003}$ & .058$_{\pm.003}$ & .037$_{\pm.001}$ & .023$_{\pm.000}$ & .499$_{\pm.029}$ & .186$_{\pm.006}$ & .092$_{\pm.002}$ & \underline{.056$_{\pm.001}$} & \underline{.037$_{\pm.001}$} \\
Greedy MI & .164$_{\pm.000}$ & \underline{.084$_{\pm.000}$} & .072$_{\pm.000}$ & .053$_{\pm.001}$ & .044$_{\pm.000}$ & .331$_{\pm.007}$ & .146$_{\pm.003}$ & \underline{.088$_{\pm.001}$} & .065$_{\pm.001}$ & .048$_{\pm.001}$ \\
\addlinespace
DPP (Gaussian($\gamma{=}0.1$)) & .574$_{\pm.001}$ & .093$_{\pm.000}$ & \underline{.051$_{\pm.000}$} & \underline{.035$_{\pm.000}$} & \underline{.023$_{\pm.000}$} & .641$_{\pm.005}$ & .195$_{\pm.007}$ & .096$_{\pm.002}$ & .060$_{\pm.001}$ & .039$_{\pm.001}$ \\
DPP (Gaussian($\gamma{=}1$)) & .574$_{\pm.001}$ & .093$_{\pm.000}$ & .051$_{\pm.000}$ & .035$_{\pm.000}$ & .023$_{\pm.000}$ & .641$_{\pm.005}$ & .200$_{\pm.006}$ & .096$_{\pm.002}$ & .061$_{\pm.001}$ & .039$_{\pm.001}$ \\
DPP (Gaussian($\gamma{=}10$)) & .574$_{\pm.001}$ & .087$_{\pm.000}$ & .051$_{\pm.000}$ & .035$_{\pm.000}$ & .024$_{\pm.000}$ & .641$_{\pm.005}$ & .176$_{\pm.006}$ & .094$_{\pm.002}$ & .062$_{\pm.001}$ & .040$_{\pm.001}$ \\
\addlinespace
DPP-MI (Gaussian($\gamma{=}0.1$)) & .164$_{\pm.000}$ & .085$_{\pm.000}$ & .071$_{\pm.000}$ & .055$_{\pm.000}$ & .046$_{\pm.000}$ & .322$_{\pm.007}$ & .150$_{\pm.003}$ & .090$_{\pm.001}$ & .066$_{\pm.001}$ & .049$_{\pm.001}$ \\
DPP-MI (Gaussian($\gamma{=}1$)) & .164$_{\pm.000}$ & .098$_{\pm.000}$ & .080$_{\pm.001}$ & .064$_{\pm.001}$ & .054$_{\pm.000}$ & \underline{.316$_{\pm.007}$} & \underline{.145$_{\pm.003}$} & .094$_{\pm.001}$ & .067$_{\pm.001}$ & .050$_{\pm.001}$ \\
DPP-MI (Gaussian($\gamma{=}10$)) & \underline{.162$_{\pm.000}$} & .105$_{\pm.001}$ & .083$_{\pm.001}$ & .072$_{\pm.001}$ & .058$_{\pm.001}$ & .319$_{\pm.007}$ & .169$_{\pm.004}$ & .104$_{\pm.002}$ & .075$_{\pm.001}$ & .051$_{\pm.001}$ \\
\midrule
FL (ref) & \textbf{.138$_{\pm.000}$} & \textbf{.073$_{\pm.000}$} & \textbf{.045$_{\pm.000}$} & \textbf{.030$_{\pm.000}$} & \textbf{.022$_{\pm.000}$} & \textbf{.306$_{\pm.005}$} & \textbf{.137$_{\pm.002}$} & \textbf{.077$_{\pm.001}$} & \textbf{.052$_{\pm.001}$} & \textbf{.036$_{\pm.001}$} \\
\bottomrule
\end{tabular}
\caption{Residual variance fraction ($\downarrow$) across budgets $k$ (20 splits). Best per column \textbf{bold}, second \underline{underlined}. facility location outperforms all the other baselines, even when the DPP variants are provided with a kernel with larger capacity.}
\label{tab:compare_rv}
\end{table*}

\label{app:benchmark_smi}

\begin{table}[t]
\centering
\footnotesize
\begin{tabular}{rl@{\hskip 1cm}rl}
\toprule
\# & Benchmark & \# & Benchmark \\
\midrule
1  & MMLU                & 19 & OJBench \\
2  & BFCL                & 20 & IF-Bench \\
3  & LiveCodeBench v6    & 21 & NexusBench \\
4  & SWE-bench (SWE-Ag.) & 22 & LiveCodeBench v5 \\
5  & $\tau^2$-Bench      & 23 & ACEBench \\
6  & IF-Eval             & 24 & HLE \\
7  & DrafterBench        & 25 & MMLU-Pro \\
8  & AIME 2025           & 26 & Multi-IF \\
9  & ComplexFuncBench    & 27 & MathOdyssey \\
10 & SimpleQA            & 28 & Gaokao 2023 \\
11 & MultiChallenge      & 29 & BFCL v4 \\
12 & InfoBench           & 30 & GPQA \\
13 & Aider Polyglot      & 31 & SWE-bench (mini) \\
14 & ComplexBench        & 32 & Math 500 \\
15 & FollowBench         & 33 & HMMT Feb25 \\
16 & OlympiadBench       & 34 & AIME 2024 \\
17 & Terminal-Bench      & 35 & $\tau$-Bench \\
18 & ToolSandbox         &    & \\
\bottomrule
\end{tabular}
\caption{Evaluation-unsupervised benchmark-level greedy ordering. Benchmarks are
ordered by the facility-location greedy on the normalized submodular mutual information
matrix $M$ (Eq.~\ref{eq:bench_smi}--\ref{eq:bench_fl}), computed purely from prompt
embeddings. Rank $i$ is the most novel benchmark
given the $i{-}1$ already selected. That is, it is important to realize that this is not a rank ordering of the individual intrinsic properties of each benchmark individually. Rather, the rank $i$ benchmark contains the next most usefulness on top of (and assuming) that we already have selected the previous $i-1$ benchmarks. For example, if we had to select only a single benchmark, MMLU would be best. If we have already selected MMLU then the next benchmark, on top of MMLU, that would be best to select is BFCL, and so on. This is all according to Figure~\ref{fig:bench_similarity_matrix}.}
\label{tab:bench_order}
\end{table}

\end{document}